\documentclass[sigconf]{acmart}
\AtBeginDocument{%
  }

\usepackage{graphicx}  
\usepackage{subfig}    

\copyrightyear{2026}
\acmYear{2026}
\setcopyright{cc}
\setcctype{by}
\acmConference[WWW '26]{Proceedings of the ACM Web Conference 2026}{April 13--17, 2026}{Dubai, United Arab Emirates}
\acmBooktitle{Proceedings of the ACM Web Conference 2026 (WWW '26), April 13--17, 2026, Dubai, United Arab Emirates}
\acmPrice{}
\acmDOI{10.1145/3774904.3792374}
\acmISBN{979-8-4007-2307-0/2026/04}

\begin{document}

\title{Enhancing Large Language Models for Time-Series Forecasting via Vector-Injected In-Context Learning}

\author{Jianqi Zhang}
\email{jluzhangjianqi@163.com}
\affiliation{%
  \institution{Institute of Software, Chinese Academy of Sciences; University of the Chinese Academy of Sciences}
  \city{Beijing}
  \country{China}
}
\authornote{Both authors contributed equally to this research.}

\author{Jingyao	Wang}
\email{jingyao_wang0728@163.com}
\affiliation{%
  \institution{Institute of Software, Chinese Academy of Sciences; University of the Chinese Academy of Sciences}
  \city{Beijing}
  \country{China}}
\authornotemark[1]

\author{Wenwen Qiang}
\email{qiang.ww0922@gmail.com}
\affiliation{%
  \institution{Institute of Software, Chinese Academy of Sciences}
  \city{Beijing}
  \country{China}}
\authornote{Corresponding author.}

\author{Fanjiang Xu}
\email{fanjiang@iscas.ac.cn}
\affiliation{%
  \institution{Institute of Software, Chinese Academy of Sciences}
  \city{Beijing}
  \country{China}}

\author{Changwen Zheng}
\email{changwen@iscas.ac.cn}
\affiliation{%
  \institution{Institute of Software, Chinese Academy of Sciences}
  \city{Beijing}
  \country{China}}

\renewcommand{\shortauthors}{Trovato et al.}

\begin{abstract}
  The World Wide Web needs reliable predictive capabilities to respond to changes in user behavior and usage patterns. Time series forecasting (TSF) is a key means to achieve this goal. In recent years, the large language models (LLMs) for TSF (LLM4TSF) have achieved good performance. However, there is a significant difference between pretraining corpora and time series data, making it hard to guarantee forecasting quality when directly applying LLMs to TSF; fine-tuning LLMs can mitigate this issue, but often incurs substantial computational overhead. Thus, LLM4TSF faces a dual challenge of prediction performance and compute overhead. To address this, we aim to explore a method for improving the forecasting performance of LLM4TSF while freezing all LLM parameters to reduce computational overhead. Inspired by in-context learning (ICL), we propose LVICL. LVICL uses our vector-injected ICL to inject example information into a frozen LLM, eliciting its in-context learning ability and thereby enhancing its performance on the example-related task (i.e., TSF). Specifically, we first use the LLM together with a learnable context vector adapter to extract a context vector from multiple examples adaptively. This vector contains compressed, example-related information. Subsequently, during the forward pass, we inject this vector into every layer of the LLM to improve forecasting performance. Compared with conventional ICL that adds examples into the prompt, our vector-injected ICL does not increase prompt length; moreover, adaptively deriving a context vector from examples suppresses components harmful to forecasting, thereby improving model performance. Extensive experiments demonstrate the effectiveness of our approach.
\end{abstract}





\maketitle

\section{Introduction}
As an ever-evolving complex system, the World Wide Web requires predictive capabilities to proactively address shifts in user behavior and usage patterns, thereby achieving efficient and stable operation \cite{kamarthi2022camul}. In this process, Time-series forecasting (TSF) plays a pivotal role \cite{ma2025mofo}. The goal of TSF is to forecast future trends and dynamics using historical data collected across the network \cite{zhang2024meta,huang2025many}. The accuracy of forecasting provides important support for the development of intelligent network services, encompassing core application scenarios such as personalized content recommendation \cite{zhang2015daily}, network economic modeling \cite{barigozzi2019nets}, and infrastructure resource allocation \cite{liu2022scinet}. In recent years, with the rapid development of large language models (LLMs), the research community has begun to explore applying LLMs to TSF. These approaches (referred to as LLM4TSF) \cite{jin2024time,liu2024autotimes,liu2025calf,chang2025llm4ts} have demonstrated strong predictive performance across multiple forecasting benchmarks.

Despite the good performance of LLM4TSF, its development still faces challenges. Specifically, one core challenge is how to mitigate the negative impact on predictive performance arising from the differences between the pre-training data (text) and the downstream data (time series) \cite{zhou2023one}. One common strategy to alleviate this issue is to fine-tune LLMs \cite{chang2025llm4ts,gururangan2020don,liu2025calf}. However, fine-tuning LLMs could incur substantial computational costs (especially in terms of GPU memory), which may limit the practical use of these methods.

To address this issue, we aim to explore an LLM4TSF approach that freezes all LLM parameters to reduce computational overhead (especially GPU memory), while, as far as possible, reproducing the predictive advantages that fine-tuning brings to LLM4TSF. Inspired by \cite{dong2022survey,zhang2023makes,li2024long}, we propose to leverage in-context learning (ICL) to achieve the above objective. ICL is a technique that leverages task-relevant examples to enhance prediction without training the LLM \cite{liu2023improving}. Specifically, ICL typically concatenates the target query with a set of related examples (question–answer pairs) into a single prompt and feeds it to the LLM to guide its response to the query \cite{zhang2023makes}. \cite{ren2024towards} shows that when tokens of examples are injected into the LLM, the forward pass is equivalent to performing an implicit gradient-descent update on the LLM. The training data for this update are the examples, and the objective of this update is to enhance the LLM’s ability to handle queries related to these examples \cite{ren2024towards}. In some cases, the effectiveness of ICL even approaches that of fine-tuning the LLM on the demonstrations \cite{dherin2025learning}. Accordingly, we argue that introducing ICL into LLM4TSF enables freezing all LLM parameters while, to the extent possible, recovering the predictive advantages that fine-tuning confers on LLM4TSF.

To verify this argument, we conduct experiments comparing the predictive performance of LLM4TSF under three settings: 1) full fine-tuning; 2) freezing the LLM and fine-tuning only the input and output layers; and 3) freezing the LLM, fine-tuning only the input and output layers, and introducing ICL (i.e., adding several examples to the prompt; see Section \ref{Sec_analysis} for details). The results in Figure \ref{fig_motivate} show that: 1) introducing ICL often reproduces part of the predictive gains achieved by fine-tuning; and 2) the magnitude of this gain is unstable and sensitive to the choice and ordering of examples, and can even be negative in some cases. Therefore, directly applying ICL to LLM4TSF may constrain the method’s performance improvement.

To address the above issues, we propose an LLM4TSF based on Vector-Injected In-Context Learning (LVICL). LVICL leverages vector-injected in-context learning to stabilize and further improve the predictive gains of applying ICL under different choices and orderings of examples. Specifically, LVICL first constructs example prompts from the training data and uses an LLM to obtain their representation vectors. Subsequently, LVICL aggregates these vectors into a single context vector in a permutation-invariant manner. This aggregation makes the context vector independent of the example order, thereby addressing the model’s sensitivity to example ordering. Next, a lightweight context adapter further refines this context vector. The adapter adaptively removes components that are detrimental to predictive performance (see Section \ref{Sec_inject} for details), thereby reducing the gain’s instability with respect to example selection and further amplifying the gain. During the model’s forward pass, the refined context vector is added to the residual stream at each layer of the LLM to assist prediction. The residual stream is a collection of intermediate representations of an LLM \cite{elhage2021mathematical}; by injecting prompt-induced features, one can write the corresponding ``memory'' into the model, thereby modulating its response tendencies \cite{panickssery2023steering} (see Section \ref{Sec_rs}). During training, the LLM’s parameters remain frozen throughout training; only the time-series input layer, output layer, and the context-refinement layer are trained.  Extensive experiments demonstrate the effectiveness of our method. Our contributions are as follows:
\begin{itemize}
    \item We investigate an issue in LLM4TSF: how to unlock the potential of LLMs for time series forecasting (TSF) with low computational costs (especially in terms of GPU memory). 
    \item  We propose LVICL, which addresses the above issue from the perspective of in-context learning (ICL). Specifically, LVICL implements ICL via vector injection, leveraging time series examples to elicit the model’s potential for TSF while keeping the LLM’s parameters frozen, thereby improving its forecasting performance on TSF.
    \item Extensive experiments on multiple real-world benchmark datasets confirm the effectiveness of the proposed method.
\end{itemize}

\section{Related Work}
\subsection{Large Language Models for Time Series}
Thanks to the rapid development of large language model (LLM) infrastructure, LLM4TSF methods have also accelerated in recent years. PromptCast \cite{xue2023promptcast} reformulates time series as textual prompts and casts forecasting as a ``sentence-to-sentence'' task; LLMTime \cite{gruver2023large} explores encoding time series as numerical tokens, demonstrating the scalability of LLMs for TSF. AutoTimes \cite{liu2024autotimes} adopts token-by-token auto-regressive prediction, aligning the learning paradigm of TSF more closely with that of LLM and thereby improving forecasting performance. FPT \cite{zhou2023one} fine-tunes LLM parameters so the model serves as a general representation extractor for time series analysis. CALF \cite{liu2025calf} uses LoRA fine-tuning to align the intermediate features of a text-LLM branch and a temporal-LLM branch, yielding better forecasting. $S^2$IP-LLM \cite{pan2024s} combines partial-parameter fine-tuning with trend–seasonality decomposition to further enhance predictive performance. LLM4TS \cite{chang2025llm4ts} proposes a two-stage fine-tuning protocol that gradually activates the language model’s capacity to model the time series modality. Unlike most prior approaches that require fine-tuning LLM, our method leverages ICL to replicate, as much as possible, the predictive benefits of fine-tuning while keeping all LLM parameters frozen, thereby reducing computational overhead (especially in GPU memory).

\subsection{In-context Learning}
\cite{brown2020language} is the first to frame in-context learning (ICL) as a simple and effective approach: adding examples to the prompt to boost LLM performance. Since then, ICL has rapidly expanded across domains, enabling LLMs to adapt to tasks ranging from traditional NLP benchmarks to more specialized settings such as kNN classification \cite{agarwal2024many} and ``jailbreaking'' \cite{anil2024many}. To improve the adaptability and effectiveness of ICL, researchers propose several enhancement strategies: 1) increasing the number of in-prompt examples to amplify ICL gains \cite{bertsch2024context,yan2023understanding,team2023gemini}; 2) using example retrieval to select the most relevant demonstrations from external corpora to boost predictive performance \cite{rubin2021learning}; and 3) exploring implicit ICL elicit the model’s predictive capabilities \cite{wangelicit,li2024implicit}. Meanwhile, recent works \cite{hojel2024finding,huang2024multimodal} extend ICL to vision and multimodal tasks and validate its effectiveness across text–vision modalities. Beyond these efforts to enhance ICL, the theory community has simultaneously conducted a series of studies aimed at explaining the principles of ICL. Multiple works \cite{ren2024towards,dherin2025learning,akyurek2022learning} show that once ICL is introduced, the forward pass of an LLM can be viewed as an implicit gradient update. This update treats the tokens of the provided examples as training data; although the objective function is expressed differently across studies, they all aim to improve the model’s performance on queries related to those examples \cite{ren2024towards}. These theoretical results suggest the plausibility of applying ICL to LLM4TSF. Inspired by the above works, this paper investigates how to efficiently apply ICL within LLM4TSF to address the challenges that LLM4TSF algorithms face.

\section{Preliminaries}
\subsection{Time Series Forecasting (TSF)} 
In TSF, we are given a historical observation sequence
\begin{equation}
X = \{\mathbf{x}_1, \ldots, \mathbf{x}_{T_h}\} \in \mathbb{R}^{T_h \times N},
\end{equation}
where $\mathbf{x}_t \in \mathbb{R}^N$ denotes the $N$ variables at time step $t$. 
The task is to predict the series of the next $T_f$ time steps:
\begin{equation}
Y = \{\mathbf{x}_{T_h+1}, \ldots, \mathbf{x}_{T_h+T_f}\} \in \mathbb{R}^{T_f \times N}.
\end{equation}
Here, $T_h$ is the look-back length, while $T_f$ is the prediction length.

\subsection{Residual Stream}
\label{Sec_rs}
\begin{figure}[htpb]
    \centering
    \includegraphics[width=0.8\linewidth]{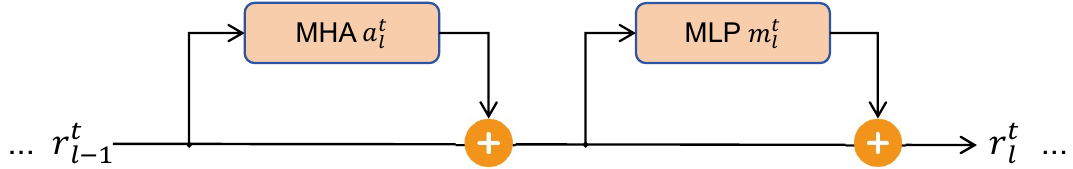}
    \caption{A schematic diagram illustrating the computation process of the $l$-th layer of a Transformer from the perspective of the residual stream. }
    \Description{}
    \label{fig_res}
\end{figure}

According to the mathematical interpretation of \cite{elhage2021mathematical}, in a Transformer, the hidden state of a single token across layers forms a residual stream. Given layer index $l$ and token position $t$, the residual stream $r_l^t$ is defined as:
\begin{equation}
\label{eq_cancha}
    r_l^t = r_{l-1}^t + a_l^t + m_l^t, 
\end{equation}
where $a_l^t$ is the output of the multi-head attention (MHA) module at layer $l$ and position $t$, and $m_l^t$ is the output of the multi-layer perceptron (MLP) at layer $l$ and position $t$. Figure \ref{fig_res} presents the MHA and MLP operations of a Transformer layer from the perspective of the residual stream, making it easier for readers to understand. As Figure \ref{fig_res} shows, $r_l^t$ is essentially the output hidden state at layer $l$ and at the $t$-th token position. The MHA and MLP can be regarded as operators that interact with the residual stream: both read from the stream and write their outputs back into it via Eq.\ref{eq_cancha}, thereby enabling the addition and deletion of information within the stream. Building on this perspective, many studies \cite{hojer2025improving,niu2024test,panickssery2023steering} directly add specific features into the residual stream to intervene in LLM behavior, steering models toward desired outcomes (e.g., ``jailbreaking'' or injecting specific “memory”). 
Inspired by these methods, we intervene in the LLM by adding the context vector (Eq.\ref{eq_eh_e}) into the residual stream, thereby making it better suited for TSF.

\begin{figure}[htbp]
    \centering
    \includegraphics[width=0.9\linewidth]{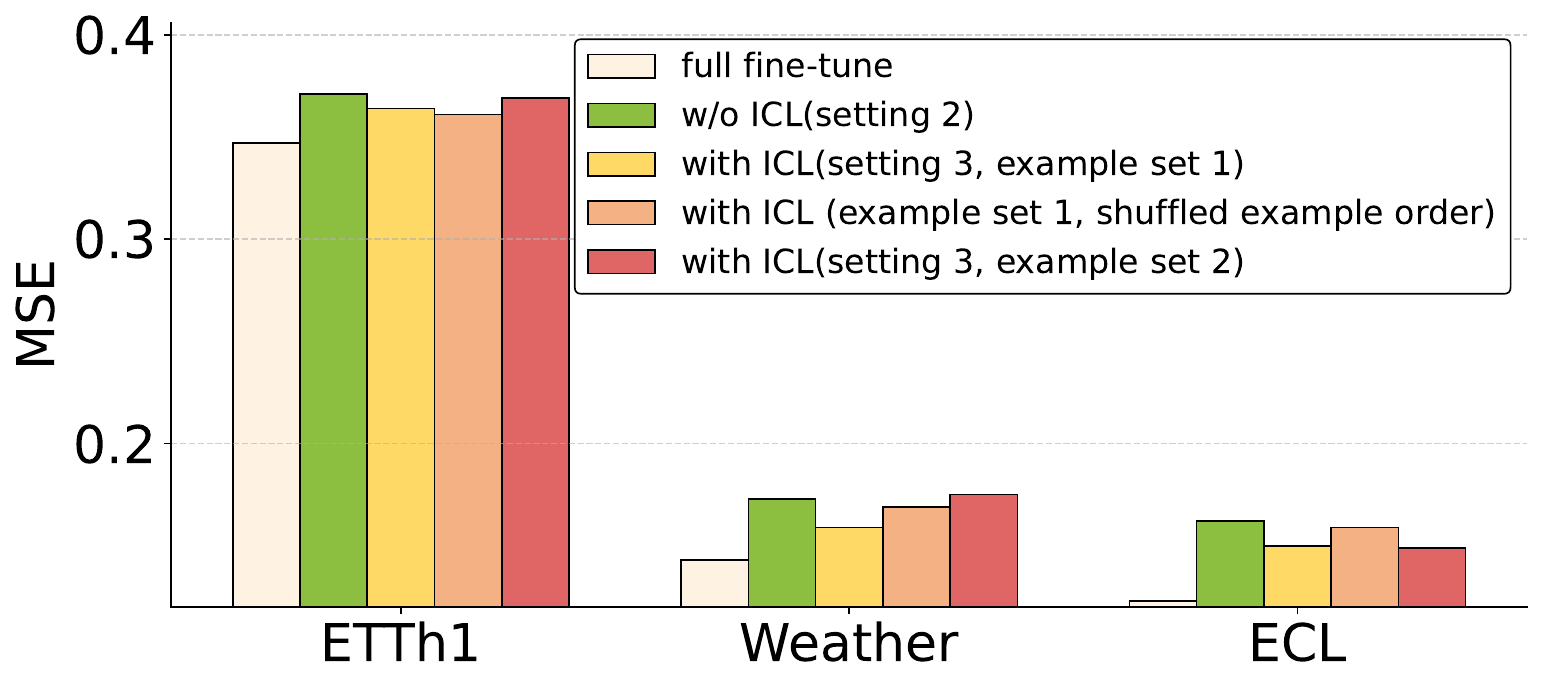}
    \caption{Comparison of forecasting performance in three settings: 1) full-parameter fine-tuning; 2) freezing the LLM and fine-tuning only the input and output layers (w/o ICL); 3) introducing ICL on top of setting 2  (with ICL).}
    \Description{}
    \label{fig_motivate}
\end{figure}

\begin{figure*}[htpb]
    \centering
    \includegraphics[width=0.80\linewidth]{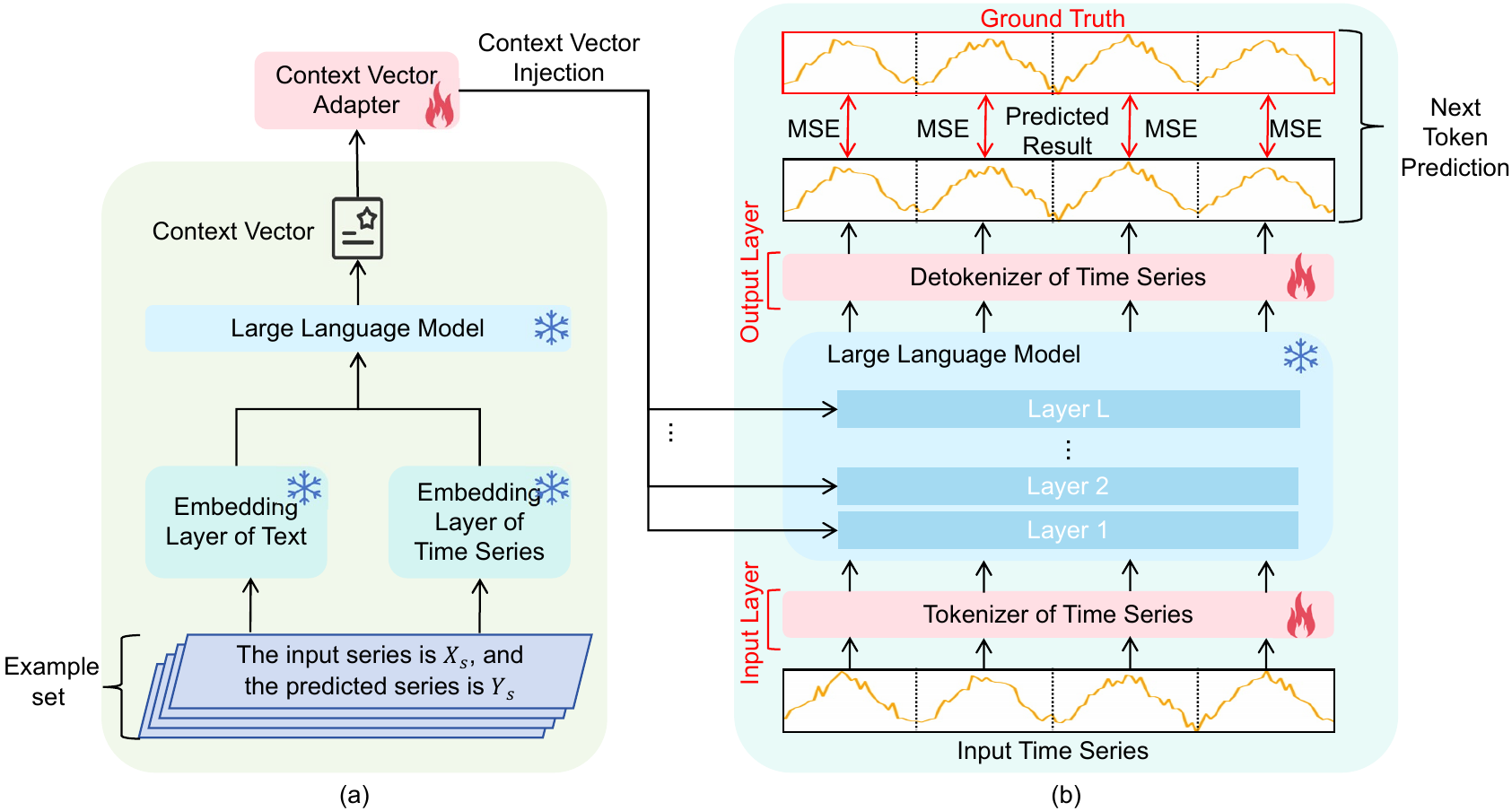}
    \caption{The overall framework of the proposed LVICL. (a) The acquisition and injection of context vectors. (b) The LLM-based time series forecasting model. }
    \Description{}
    \label{fig_method}
\end{figure*}

\section{Empirical Analysis}
\label{Sec_analysis}
In this subsection, we empirically verify that introducing ICL into LLM4TSF enables freezing all LLM parameters while recovering the predictive advantages that fine-tuning confers on LLM4TSF.

We first present the LLM4TSF configuration used in our experiments, then explain how ICL is incorporated into LLM4TSF, and finally describe the experimental procedure. Firstly, the LLM4TSF employs a simple ``input layer + LLM + output layer'' architecture. We use Llama-2-7B \cite{touvron2023llama} as the LLM. In our experiments, the look-back length is set to $672$ and the forecast length is set to $96$. On the input side, the input series $X_{\text{in}}$ is divided into non-overlapping patches of length $96$. The resulting sequence of patches is fed into the model to generate the forecast series. The model makes predictions in an autoregressive manner. Secondly, we describe how ICL is introduced for LLM4TSF. Specifically, we randomly select a fixed set of $100$ training samples ${(X_s^i, Y_s^i)}_{i=1}^{i=100}$, where each $X_s^i$ contains $672$ time steps and each $Y_s^i$ contains $96$ time steps. By concatenating these examples and then appending the input series at the end, we form the final input prompt. Next, the above prompt is input into the model to produce the forecast series. Following the above steps, we incorporate ICL into LLM4TSF. Thirdly, we describe the experimental procedure. Specifically, we evaluate the forecasting performance of the above LLM4TSF under three settings: 1) full-parameter fine-tuning; 2) freezing the LLM and fine-tuning only the input and output layers; 3) introducing ICL on top of setting 2. The experiments with ICL (setting 3) are conducted under three example setups: the first two use the first example set but in different orders, and the third uses the second example set. The results are shown in Figure \ref{fig_motivate}. To prevent bias caused by random factors, all experimental results are averaged over five random seeds.

The results in Figure \ref{fig_motivate} show that: 1) introducing ICL often reproduces part of the predictive gains achieved by fine-tuning; and 2) the magnitude of this gain is unstable and sensitive to the choice and ordering of examples, and can even be negative in some cases. Therefore, directly applying ICL to LLM4TSF may constrain the method’s performance improvement.

\section{Methodology}
In this section, we propose a Large Language Model for Time-Series Forecasting based on Vector-Injected In-Context Learning (LVICL). LVICL leverages vector-injected in-context learning to stabilize and further improve the predictive gains of applying ICL under different choices and orderings of examples. Specifically, we first describe the overall pipeline of LVICL (Section \ref{sec_pipeline}). Next, we explain how to obtain the context vector from examples (Section \ref{Sec_acqu}). Then, we describe how to inject the context vector into the model to perform in-context learning (Section \ref{Sec_inject}). Finally, we demonstrate the optimization procedure of LVICL (Section \ref{Sec_opt}).

\subsection{Overall Pipeline}
\label{sec_pipeline}
In this subsection, we present the overall pipeline of our LVICL. Figure \ref{fig_method} illustrates its architecture. Specifically, as shown at the bottom of Figure \ref{fig_method}(a), we first convert the time series data into time-series examples. For each example, we then use a fixed LLM to transform it into a representation vector. These representation vectors are subsequently aggregated into a single context vector in a permutation-invariant manner. This vector carries a compressed representation of all examples and can be directly added to the LLM’s residual stream to inject example information. Compared with adding examples to the prompt, this injection method incurs lower computational overhead. During the forward process, these context vectors are further refined by a context vector adapter. The adapter adaptively suppresses prediction-harming components of the context vector by weighting its different channels, thereby improving the performance of ICL. With these vectors in hand, we proceed to process the input time series. As shown at the bottom of Figure \ref{fig_method}(b), we partition the input multivariate time series into multiple univariate series and handle them separately. Next, for each series, we split it into non-overlapping patches of length $P$. Each segment is mapped to a token by a tokenizer in the input layer. The tokenizer is a fully connected layer. Then, the resulting token sequence is fed into the LLM for further modeling. While the LLM processes tokens, the context vector computed at the top of Figure \ref{fig_method}(a) is injected into each layer of the LLM. Finally, the LLM’s output is decoded by a single-layer fully connected layer (i.e., the output layer). As the top of Figure \ref{fig_method}(b) shows, decoding adopts an autoregressive prediction strategy.

\subsection{Context Vector Acquisition}
\label{Sec_acqu}
In this subsection, we detail how to construct time series examples and derive a context vector from them. The vector is then injected into the LLM to enable ICL. This section corresponds to the lower-middle part of Fig. \ref{fig_method}(a). Specifically, the procedure comprises three steps: 1) generate textual examples for each time series; 2) extract representation vectors from these texts; and 3) aggregate the vectors from multiple examples to obtain the context vector.

Firstly, we describe how to generate the textual example for each time series. Specifically, each time series includes the historical series $X_s$ and the corresponding future series $Y_s$. For each time series, we use a simple natural language description as the example: 
\begin{equation}
    \label{eq_example}
    The\ input\ series\ is\ X_s,\ and\ the\ predicted\ series\ is\ Y_s
\end{equation}
The examples corresponding to all time series form a set (as shown at the bottom of Figure \ref{fig_method}(a)). The set contains multiple input–output pairs for the specific task (i.e., TSF), enabling the LLM to learn patterns for handling the task \cite{bertsch2024context}. Subsequent modules will extract representations from this set to enhance the LLM’s predictive performance in TSF.

Secondly, we describe how to extract representation vectors from the example in Eq.\ref{eq_example}. According to prior ICL work in NLP \cite{he2021efficient,wangelicit}, we can obtain the representation vector for each example by feeding it into the LLM. However, the LLM only includes an embedding layer for text (denoted $\mathrm{Emb}_n$) and lacks one for the time series data $X_s$ and $Y_s$ (denoted $\mathrm{Emb}_t$). To address this issue, we propose to get $\mathrm{Emb}_t$ through training. Specifically, we first construct a simple LLM4TSF whose architecture is the same as LVICL but with all context-vector–related modules removed (i.e., the architecture in Figure \ref{fig_method}(b)). Then, we use the training data of each dataset to train the input/output layers with the LLM frozen. After training, we take the input layer as the dataset’s $\mathrm{Emb}_t$. Next, we use $\mathrm{Emb}_n$ and $\mathrm{Emb}_t$ to encode the textual and time series parts of Eq.\ref{eq_example}, respectively. The token features for the example in Eq.\ref{eq_example} are as follows:
\begin{equation}
\label{eq_example_t}
\operatorname{cat}\!\left[
\begin{array}{l}
\mathrm{Emb}_n\!\big(\text{The input series is}\big),
\mathrm{Emb}_t\!\big(X_s\big),\\
\mathrm{Emb}_n\!\big(\text{, and the predicted series is}\big),
\mathrm{Emb}_t\!\big(Y_s\big)
\end{array}
\right]
\end{equation}
Here, $cat$ is the operator that concatenates token sequences. Finally, we feed the token sequences from Eq.\ref{eq_example_t} into the LLM to obtain the representation vectors. According to the autoregressive property \cite{Transformer}, only the final token can aggregate information from all preceding tokens. Consequently, the hidden representation of the final token best characterizes the context. Based on this, we propose to use the representation of the last token as the representation vector. Formally, the representation vector of an example is:
\begin{equation}
    \label{eq_eh}
    V_i=\{r^e_{l,i}\}^{l=1,...,L}\ \ i=1,...,N,
\end{equation}
where $r^e_{l, i}$ denotes the representation of the last output token of layer $l$ when the LLM processes the $i$-th sample, the superscript $e$ denotes the last token, $N$ is the total number of examples, $L$ is the number of LLM layers.

Thirdly, we aggregate the vectors from multiple examples to obtain the final context vector. Specifically, we propose to aggregate the representation vectors of all samples by averaging them to form the final context vector:
\begin{equation}
    \label{eq_eh_e}
    V_{final}=\{\mathbb{E} _i[r^e_{l,i}]\}^{l=1,...,L}. 
\end{equation}
By taking the mean, we can accentuate features common to multiple examples while suppressing example-specific components. \cite{todd2023function} shows that these common features carry more task-specific information. In addition, mean aggregation reduces the sensitivity of the gain to sample ordering, thereby improving stability.

Through the three steps above, we obtain a context vector that can be injected into the LLM. Compared with examples in traditional ICL, on the one hand, the context vector can be injected into the model more quickly (see Section \ref{Sec_inject} for injection details); On the other hand, since the context vector is built via a permutation-invariant aggregation, it exhibits no sensitivity to the order of examples, thereby effectively mitigating the traditional ICL’s sensitivity to example ordering. These three steps can be completed before training and testing, and the resulting context vector can be reused in subsequent processes.

\subsection{Context Vector Injection}
\label{Sec_inject}
In this section, we describe how to inject the context vector into the model. The process has two steps: 1) further refine the context vector; and 2) add the refined vector to the residual stream. Specifically, we propose to refine the context vector, as the vector obtained in the previous subsection still contains components that are detrimental to prediction. These components mainly manifest in two ways: first, some channels of the context vector have overly large magnitudes, which would overwhelm small-magnitude values in the residual stream and thus hinder prediction; second, the examples may contain factors unrelated to TSF, and these factors are propagated into the context vector. To adaptively suppress the above components, we propose using a simple context vector adapter $f_{adp}$ to post-process the context vector $V_{final}$:
\begin{equation}
    \label{eq_eh_e1}
    V_{final}^l=f_{adp}(\mathbb{E} _i[r^e_{l,i}]),\ l=1,...,L.
\end{equation}
where $f_{adp}$ is implemented with a fully connected layer. Then, we inject example-related information into the model by adding the refined context vector to the residual stream (details in Sec. \ref{Sec_rs}):
\begin{equation}
    \label{eq_context_inject}
    r_{l}^t=r_{l-1}^t + a_l^t + m_l^t+V_{final}^l,
\end{equation}
where $l=1,...,L,$, $ t=1,...,T$, $T$ is the number of tokens.

In summary, we use a context vector adapter to suppress the distracting components of the context vector, thereby reducing the ICL gain’s instability with respect to example selection and further amplifying the gain; we then inject the refined vector into the residual stream via element-wise addition to perform the vector-injected ICL, thereby improving TSF prediction performance. This injection procedure requires only a single fully connected layer and one element-wise addition. Our experiments in Appendix \ref{sec_IE_ICL} and Section \ref{sec_Sensitivity} show that, compared with traditional ICL, which concatenates examples into the prompt, our method incurs lower computational overhead and achieves better predictive performance.

\subsection{Model Optimization}
\label{Sec_opt}
In this subsection, we introduce the optimization procedure of LVICL. The trainable parameters include the input layer (denoted by $\theta_i$) and output layer (denoted by $\theta_o$) in Fig.\ref{fig_method}(b), as well as the context vector adapter (denoted by $\theta_a$) in Fig.\ref{fig_method}(a). The optimization objective of the model is:
\begin{equation}
    \label{eq_optimization}
    \mathop {\min }\limits_{\theta_i,\theta_o,\theta_a}{\mathcal{L}_{\text{mse}}}({{f}_{\theta_i,\theta_o,\theta_a,\theta_{LLM}} }(X),Y),
\end{equation}
where $\theta_{LLM}$ is the parameter of LLM, $\mathcal{L}_{\text{mse}}$ is the mean squared error (MSE) loss. With these optimizations, the model minimizes end-to-end error while updating only a few parameters.

\section{Experiments}
This section empirically validates our LVICL. Specifically, we first describe the experimental setup and then present comparative results on multiple datasets. Next, we demonstrate the effectiveness of LVICL through ablation studies and additional analyses. The hyperparameter sensitivity experiments are in Appendix \ref{sec_hsa}.
\subsection{Experimental Settings}
\subsubsection{Datasets}
For long-term time series forecasting, we use seven real-world datasets (ECL, ETT (h1, h2, m1, m2), Traffic, and Weather). For short-term forecasting, we adopt the M4 dataset, which is one of the most commonly used in LLM4TSF methods \cite{liu2024autotimes,liu2025calf,jin2024time}. For zero-shot experiments, we follow AutoTimes \cite{liu2024autotimes} and use the M3 and M4 datasets. The train–validation–test splits follow \cite{liu2023itransformer} and \cite{liu2024autotimes}. Further details on the datasets are in Appendix \ref{Sec_data_des}.

\subsubsection{Baselines}
We evaluate a broad set of representative time-series forecasting (TSF) methods, including 1) LLM4TSF methods such as AutoTimes \cite{liu2024autotimes}, FPT \cite{zhou2023one}, TIME-LLM \cite{jin2024time}, and UniTime \cite{liu2024unitime}; 2)other deep learning baselines such as DLinear \cite{DLinear}, TimesNet \cite{Timesnet}, PatchTST \cite{PatchTST}, Fedformer \cite{fedformer}, and iTransformer \cite{liu2023itransformer}.

\subsubsection{Implementation Details:}
All experiments are conducted on NVIDIA A100-80G GPUs using PyTorch \cite{imambi2021pytorch}. We adopt the Adam optimizer \cite{kingma2014adam} with initial learning rates selected from \{3e-5, 5e-5, 7e-5, 9e-5\}, and use MSE loss for optimization. Our method makes predictions in an autoregressive manner and transforms the multivariate data into univariate data by treating each feature of the sequence as an individual time series. The setting of the length of the patch $P$ can be found in Table \ref{tab_patch_l}. The look-back length for ETT, Weather, ECL, and Traffic is set to 672. The look-back length for M3 and M4 is set to twice the forecast length. The forecast lengths for each dataset are listed in Table \ref{tab_data}. We adopt LLaMA-7B \cite{touvron2023llama} as our base LLM, and we also conduct additional experiments with other LLMs to validate the effectiveness of our approach (Section \ref{sec_gen}). The batch size is chosen from \{192, 256, 1024, 2048\}, and the number of training epochs is capped at 40. Early stopping is triggered if the validation loss fails to decrease for three consecutive evaluations. For a single example $(X_s, Y_s)$ in Eq.\ref{eq_example}, the length of $X_s$ is the same as the look-back length of each dataset, and the length of $Y_s$ is the same as the minimum predicted length of each dataset. For Traffic and ECL, which have a large amount of training data, we randomly select 5\% of the training data to generate context vectors. For other datasets, we use all the training data to generate context vectors. The context vector can be generated before training and does not affect the model’s subsequent inference speed.

\subsection{Standard Dataset Evaluation}
To validate the effectiveness of our approach, we compare it with other methods under both long-term and short-term forecasting settings. The experimental results are shown in Table \ref{tab_long_res_avg} and Table \ref{tab_short_res}. Our results are obtained by averaging over five runs with different random seeds. In the tables, the best results are highlighted in bold, and the second-best are underlined. From the results, it can be seen that LVICL demonstrates strong competitiveness in both settings. Unlike most LLM4TSF methods that require modality alignment to boost performance, our method achieves better forecasting without any modality alignment, further confirming its effectiveness. 

\begin{table*}[t]
\caption{Long-term forecasting results averaged from four prediction lengths $\in$ \{96, 192, 336, 720\}. Detailed results can be found in Appendix \ref{full_res}. Most results of other methods derive from \cite{liu2024autotimes}. Results not reported in \cite{liu2024autotimes} are obtained by running each method’s source code; the look-back window is selected from \{36, 96, 192, 336, 512, 672, 720\} to report the best results.}
\centering
\small
\setlength{\tabcolsep}{1.5pt}
\resizebox{0.9\textwidth}{!}{%
\begin{tabular}{l|rr|rr|rr|rr|rr|rr|rr|rr|rr|rr|rr}
\toprule
\textbf{Dataset} & \multicolumn{2}{c}{\textbf{LVICL}} & \multicolumn{2}{c}{AutoTimes} & \multicolumn{2}{c}{TimeLLM} & \multicolumn{2}{c}{FPT} & \multicolumn{2}{c}{Unitime} & \multicolumn{2}{c}{iTransformer} & \multicolumn{2}{c}{DLinear} & \multicolumn{2}{c}{PatchTST} & \multicolumn{2}{c}{TimesNet} & \multicolumn{2}{c}{Chronos} & \multicolumn{2}{c}{TimeFM} \\
 & \textbf{MSE} & \textbf{MAE} & \textbf{MSE} & \textbf{MAE} & \textbf{MSE} & \textbf{MAE} & \textbf{MSE} & \textbf{MAE} & \textbf{MSE} & \textbf{MAE} & \textbf{MSE} & \textbf{MAE} & \textbf{MSE} & \textbf{MAE} & \textbf{MSE} & \textbf{MAE} & \textbf{MSE} & \textbf{MAE} & \textbf{MSE} & \textbf{MAE} & \textbf{MSE} & \textbf{MAE} \\
\midrule
\textbf{ETTh1} & \textbf{0.381} & \textbf{0.412} & \underline{0.389} & \underline{0.422} & 0.408 & 0.423 & 0.427 & 0.426 & 0.442 & 0.448 & 0.438 & 0.450 & 0.423 & 0.437 & 0.413 & 0.431 & 0.458 & 0.450 & 0.401 & 0.434 & 0.417 & 0.451 \\
\textbf{ETTh2} & \textbf{0.326} & \textbf{0.376} & 0.352 & 0.395 & 0.354 & 0.393 & 0.353 & 0.391 & 0.356 & 0.399 & 0.382 & 0.414 & 0.431 & 0.446 & 0.330 & 0.379 & 0.414 & 0.427 & \underline{0.327} & \underline{0.377} & 0.334 & 0.385 \\
\textbf{ETTm1} & \underline{0.328} & \underline{0.378} & 0.332 & 0.380 & 0.350 & \underline{0.378} & 0.366 & 0.382 & 0.375 & 0.403 & 0.370 & 0.399 & 0.357 & \underline{0.378} & 0.351 & 0.380 & 0.400 & 0.406 & \textbf{0.322} & \textbf{0.371} & 0.341 & 0.393 \\
\textbf{ETTm2} & \textbf{0.239} & \textbf{0.306} & 0.243 & 0.310 & 0.254 & 0.314 & 0.265 & 0.315 & 0.277 & 0.329 & 0.272 & 0.331 & 0.267 & 0.333 & 0.255 & 0.315 & 0.291 & 0.333 & 0.245 & 0.313 & \underline{0.242} & \underline{0.309} \\
\textbf{ECL} & \textbf{0.158} & \textbf{0.248} & \underline{0.159} & \underline{0.253} & \underline{0.159} & \underline{0.253} & 0.167 & 0.263 & 0.216 & 0.305 & 0.161 & 0.256 & 0.177 & 0.274 & \underline{0.159} & \underline{0.253} & 0.192 & 0.295 & 0.164 & 0.258 & 0.162 & 0.255 \\
\textbf{Weather} & \underline{0.224} & \underline{0.262} & 0.235 & 0.273 & 0.225 & \textbf{0.257} & 0.237 & 0.270 & 0.253 & 0.276 & 0.238 & 0.272 & 0.240 & 0.300 & 0.226 & 0.264 & 0.259 & 0.287 & \textbf{0.220} & 0.263 & 0.229 & 0.270 \\
\textbf{Traffic} & \textbf{0.370} & \textbf{0.260} & \underline{0.374} & \underline{0.264} & 0.388 & \underline{0.264} & 0.414 & 0.294 & 0.429 & 0.304 & 0.379 & 0.272 & 0.434 & 0.295 & 0.391 & \underline{0.264} & 0.620 & 0.336 & 0.376 & \underline{0.264} & 0.383 & 0.269 \\
\bottomrule
\end{tabular}
}
\label{tab_long_res_avg}
\end{table*}

\begin{table*}[t]
\centering
\small
\setlength{\tabcolsep}{5pt}
\caption{Full results for the short-term forecasting task. All results of other methods derive from \cite{liu2024autotimes}.}
\begin{tabular}{llccccccc}
\toprule
\textbf{Scenario} & \textbf{Category} & \textbf{LVICL} & AutoTimes & TimeLLM & FPT & DLinear & PatchTST & TimesNet \\
\midrule
{Year} & SMAPE & \textbf{13.101} & \underline{13.319} & 13.419 & 13.531 & 13.866 & 13.517 & 13.394 \\
 & MASE & \textbf{2.988} & \underline{2.993} & 3.005 & 3.015 & 3.006 & 3.031 & 3.004 \\
 & OWA & \textbf{0.775} & \underline{0.784} & 0.789 & 0.793 & 0.802 & 0.795 & 0.787 \\
\midrule
{Quarter} & SMAPE & \textbf{10.010} & \underline{10.101} & 10.110 & 10.177 & 10.689 & 10.847 & \underline{10.101} \\
 & MASE & \textbf{1.170} & 1.182 & \underline{1.178} & 1.194 & 1.294 & 1.315 & 1.183 \\
 & OWA & \textbf{0.882} & 0.890 & \underline{0.889} & 0.897 & 0.957 & 0.972 & 0.890 \\
\midrule
{Month} & SMAPE & \textbf{12.321} & \underline{12.710} & 12.980 & 12.894 & 13.372 & 14.584 & 12.866 \\
 & MASE & \textbf{0.934} & \textbf{0.934} & 0.963 & \underline{0.956} & 1.014 & 1.169 & 0.964 \\
 & OWA & \textbf{0.867} & \underline{0.880} & 0.903 & 0.897 & 0.940 & 1.055 & 0.894 \\
\midrule
{Others} & SMAPE & \textbf{4.751} & 4.843 & \underline{4.795} & 4.940 & 4.894 & 6.184 & 4.982 \\
 & MASE & \textbf{3.127} & 3.277 & \underline{3.178} & 3.228 & 3.358 & 4.818 & 3.323 \\
 & OWA & \textbf{0.994} & 1.026 & \underline{1.006} & 1.029 & 1.044 & 1.140 & 1.048 \\
\midrule
{AVERAGE} & SMAPE & \textbf{11.567} & \underline{11.831} & 11.983 & 11.991 & 12.418 & 13.022 & 11.930 \\
 & MASE & \textbf{1.573} & \underline{1.585} & 1.595 & 1.600 & 1.656 & 1.814 & 1.597 \\
 & OWA & \textbf{0.838} & \underline{0.850} & 0.859 & 0.861 & 0.891 & 0.954 & 0.867 \\
\bottomrule
\end{tabular}
\label{tab_short_res}
\end{table*}

\begin{table*}[t]
\caption{Full results for the zero-shot task. All results of other methods derive from \cite{liu2024autotimes}. SMAPE is adopted as the metric.}
\centering
\resizebox{0.70\linewidth}{!}{%
\small
\setlength{\tabcolsep}{7pt}
\renewcommand{\arraystretch}{1.15}
\begin{tabular}{llccccccc}
\toprule
\textbf{Scenario} & \textbf{Category} & \textbf{LVICL} & AutoTimes & FPT & DLinear & PatchTST & TimesNet & FEDformer \\
\midrule
M4→M3 & Year & \textbf{15.11} & \underline{15.71} & 16.42 & 17.43 & 15.99 & 18.75 & 16.00 \\
 & Quarter & \underline{9.45} & \textbf{9.35} & 10.13 & 9.74 & 9.62 & 12.26 & 9.48 \\
 & Month & \textbf{13.98} & 14.06 & 14.10 & 15.65 & 14.71 & \underline{14.01} & 15.12 \\
 & Others & \underline{5.59} & 5.79 & \textbf{4.81} & 6.81 & 9.44 & 6.88 & 8.94 \\
 & Average & \textbf{12.596} & \underline{12.75} & 13.06 & 14.03 & 13.39 & 14.17 & 13.53 \\
\midrule
M3→M4 & Year & \textbf{13.721} & \underline{13.728} & 13.740 & 14.193 & 13.966 & 15.655 & 13.887 \\
 & Quarter & \textbf{10.734} & \underline{10.742} & 10.787 & 18.856 & 10.929 & 11.877 & 11.513 \\
 & Month & \underline{14.560} & \textbf{14.558} & 14.63 & 14.765 & 14.664 & 16.165 & 18.154 \\
 & Others & \textbf{6.219} & \underline{6.259} & 7.081 & 9.194 & 7.087 & 6.863 & 7.529 \\
 & Average & \textbf{13.032} & \underline{13.036} & 13.125 & 15.337 & 13.228 & 14.553 & 15.047 \\
\bottomrule
\end{tabular}
}  
\label{tab:tsf_no_multirow}
\end{table*}

\begin{figure}[htbp]
    \centering
    \includegraphics[width=0.85\linewidth]{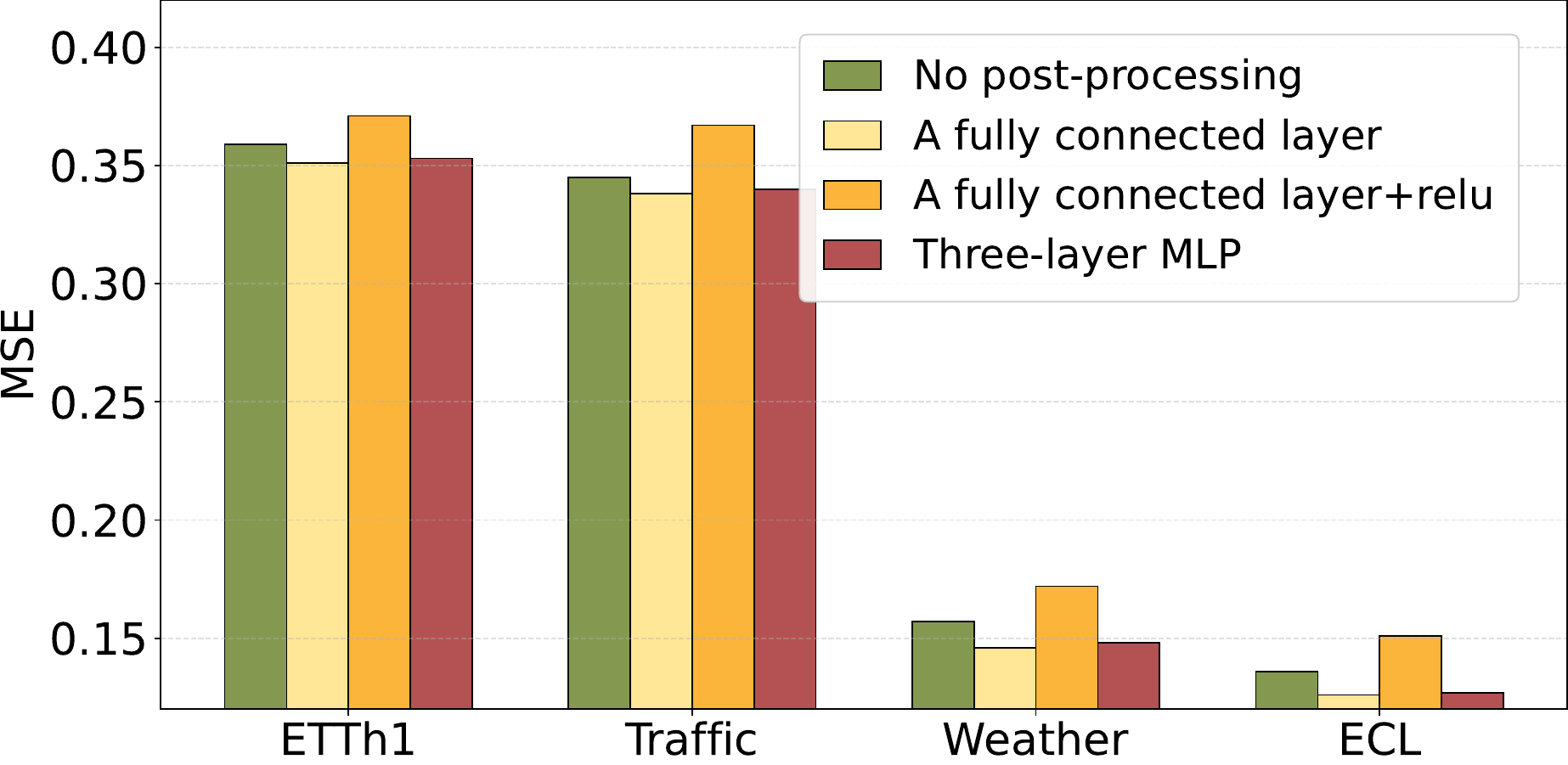}
    \caption{Impact of post-processing strategies for context vectors.}
    \Description{}
    \label{fig_mlp}
\end{figure}

\begin{figure}[htbp]
    \centering
    \includegraphics[width=0.85\linewidth]{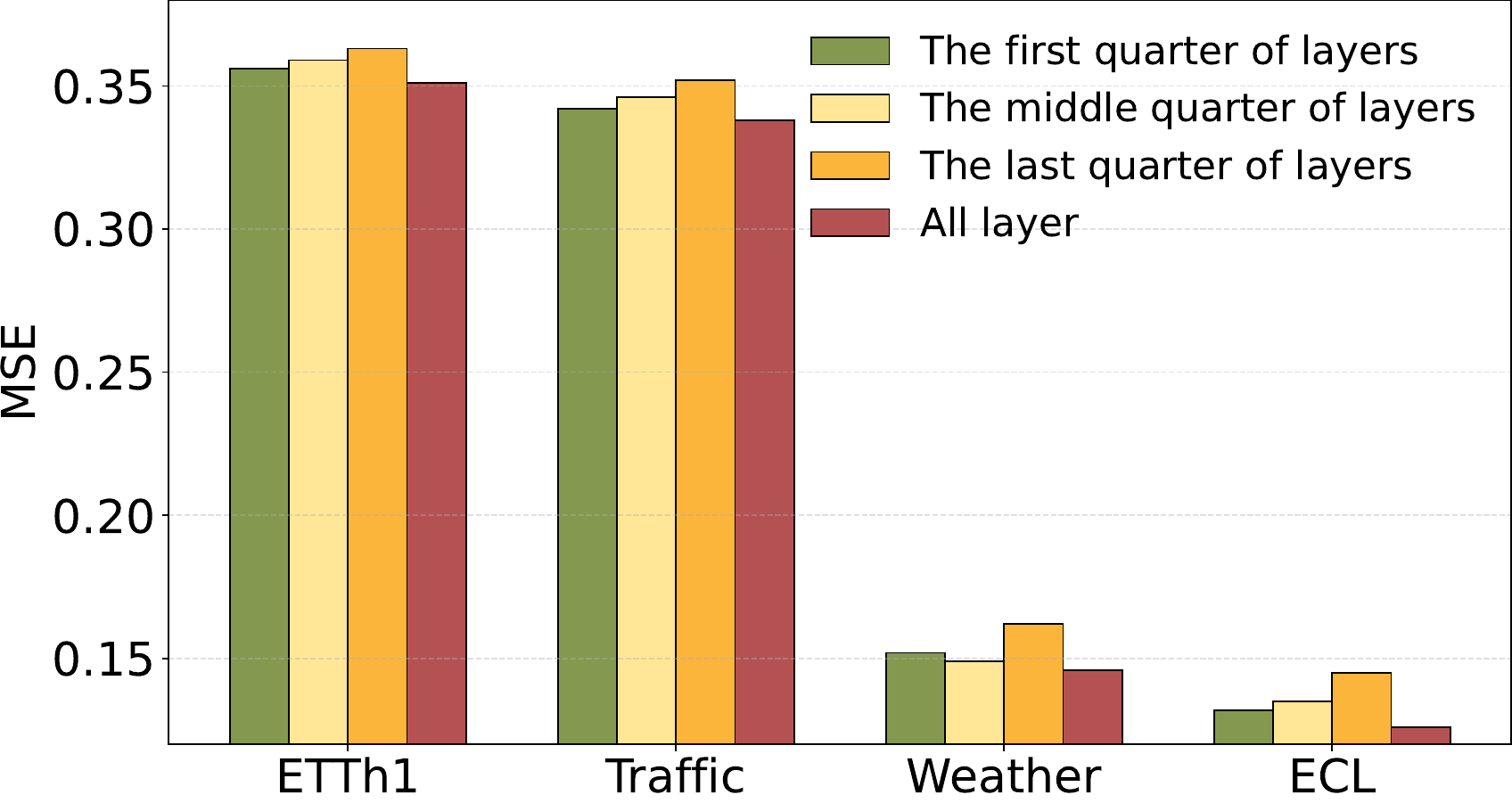}
    \caption{Impact of context vector injection location. }
    \Description{}
    \label{fig_insert}
\end{figure}

\begin{figure}[htbp]
    \centering
    \includegraphics[width=0.9\linewidth]{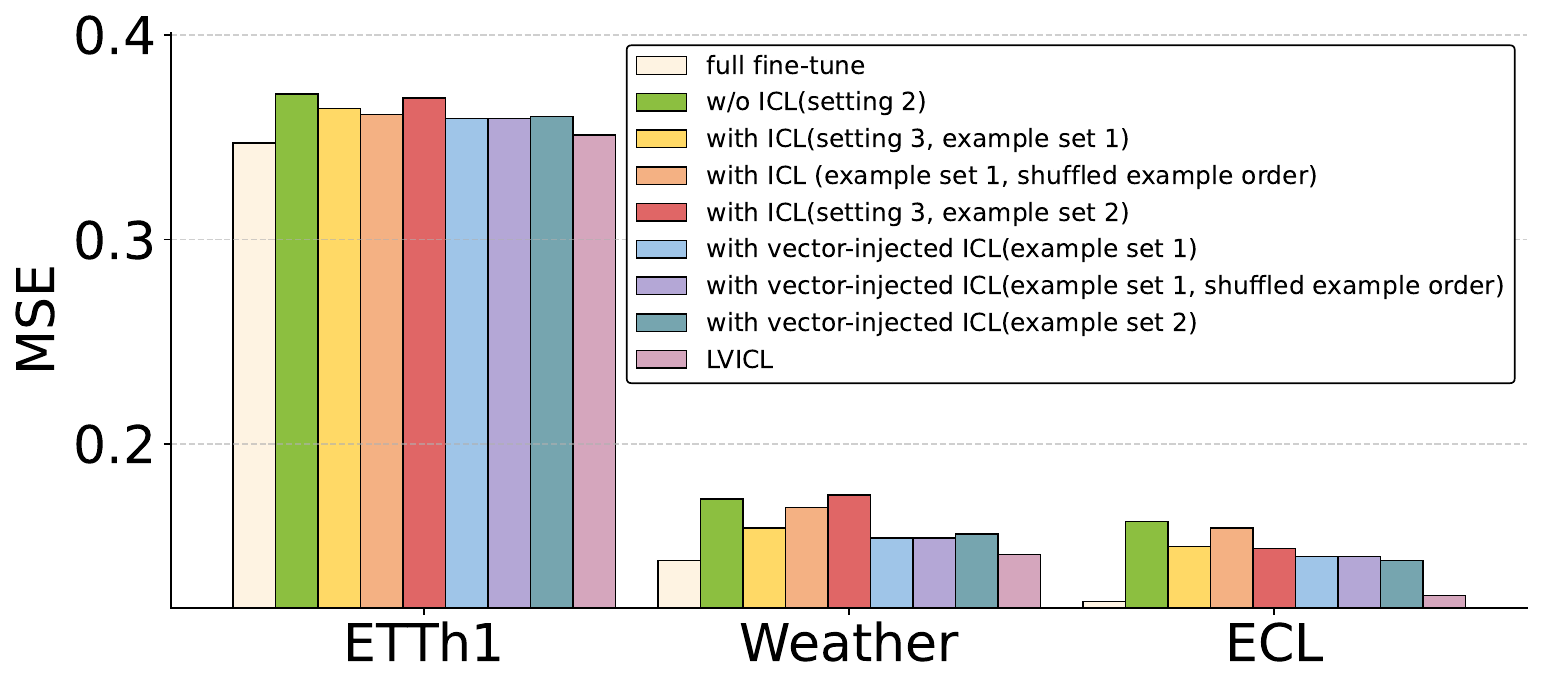}
    \caption{Sensitivity experiments on the choice and ordering of examples using vector-injected ICL and traditional ICL. }
    \Description{}
    \label{fig_sense}
\end{figure}

\begin{table*}[ht]
\caption{Results of alternative LLMs.}
\centering
\resizebox{0.72\linewidth}{!}{%
\footnotesize
\setlength{\tabcolsep}{9pt}
\begin{tabular}{ll|cc|cc|cc|cc|cc}
\toprule
Dataset & Prediction Length & \multicolumn{2}{c|}{GPT-2 (124M)} & \multicolumn{2}{c|}{OPT-350M} & \multicolumn{2}{c|}{OPT-1.3B} & \multicolumn{2}{c|}{OPT-2.7B} & \multicolumn{2}{c}{LLAMA-7B} \\
 &  & MSE & MAE & MSE & MAE & MSE & MAE & MSE & MAE & MSE & MAE \\
\midrule
ETTh1 & 96 & 0.355 & \textbf{0.382} & 0.356 & 0.392 & \textbf{0.348} & 0.384 & 0.351 & 0.387 & 0.351 & 0.389 \\
 & 192 & 0.382 & 0.408 & 0.386 & 0.412 & 0.380 & \textbf{0.406} & 0.380 & 0.408 & \textbf{0.379} & 0.408 \\
 & 336 & 0.399 & 0.418 & 0.402 & 0.422 & 0.399 & 0.419 & 0.395 & 0.422 & \textbf{0.392} & \textbf{0.417} \\
 & 720 & 0.425 & 0.446 & 0.428 & 0.451 & 0.426 & 0.446 & 0.417 & 0.441 & \textbf{0.402} & \textbf{0.434} \\
 & Avg & 0.390 & 0.414 & 0.393 & 0.419 & 0.388 & 0.414 & 0.386 & 0.414 & \textbf{0.381} & \textbf{0.412} \\
\midrule
ECL & 96 & 0.137 & 0.235 & 0.133 & 0.232 & 0.129 & 0.227 & 0.129 & 0.226 & \textbf{0.126} & \textbf{0.224} \\
 & 192 & 0.160 & 0.247 & 0.155 & 0.243 & 0.151 & 0.239 & 0.150 & 0.238 & \textbf{0.148} & \textbf{0.235} \\
 & 336 & 0.176 & 0.264 & 0.170 & 0.261 & 0.166 & 0.256 & 0.166 & 0.256 & \textbf{0.161} & \textbf{0.252} \\
 & 720 & 0.213 & 0.297 & 0.208 & 0.295 & 0.203 & 0.290 & 0.204 & 0.291 & \textbf{0.196} & \textbf{0.282} \\
 & Avg & 0.172 & 0.261 & 0.167 & 0.258 & 0.162 & 0.253 & 0.162 & 0.253 & \textbf{0.158} & \textbf{0.248} \\
\midrule
Traffic & 96 & 0.364 & 0.254 & 0.366 & 0.257 & 0.356 & 0.250 & 0.353 & 0.248 & \textbf{0.338} & \textbf{0.245} \\
 & 192 & 0.390 & 0.272 & 0.388 & 0.276 & 0.381 & 0.267 & 0.377 & 0.265 & \textbf{0.360} & \textbf{0.251} \\
 & 336 & 0.416 & 0.271 & 0.414 & 0.272 & 0.405 & 0.266 & 0.402 & 0.264 & \textbf{0.382} & \textbf{0.259} \\
 & 720 & 0.436 & 0.300 & 0.433 & 0.301 & 0.427 & 0.296 & 0.424 & 0.294 & \textbf{0.401} & \textbf{0.285} \\
 & Avg & 0.401 & 0.274 & 0.400 & 0.277 & 0.392 & 0.270 & 0.389 & 0.268 & \textbf{0.370} & \textbf{0.260} \\
\midrule
Weather & 96 & 0.151 & 0.201 & 0.150 & 0.200 & 0.150 & 0.200 & 0.149 & 0.198 & \textbf{0.146} & \textbf{0.196} \\
 & 192 & 0.191 & 0.238 & 0.189 & 0.237 & \textbf{0.188} & \textbf{0.236} & 0.190 & 0.239 & 0.190 & 0.240 \\
 & 336 & 0.246 & 0.285 & 0.245 & 0.281 & 0.247 & 0.283 & 0.248 & 0.284 & \textbf{0.240} & \textbf{0.280} \\
 & 720 & 0.337 & 0.365 & 0.330 & 0.354 & 0.329 & 0.359 & 0.346 & 0.366 & \textbf{0.320} & \textbf{0.340} \\
 & Avg & 0.231 & 0.272 & 0.228 & 0.268 & 0.228 & 0.269 & 0.233 & 0.272 & \textbf{0.224} & \textbf{0.264} \\
\bottomrule
\end{tabular}
}  
\label{tab_gen_res}
\end{table*}


\begin{figure*}[htbp]
    \centering
    \subfloat[]{\includegraphics[width=0.30\textwidth]{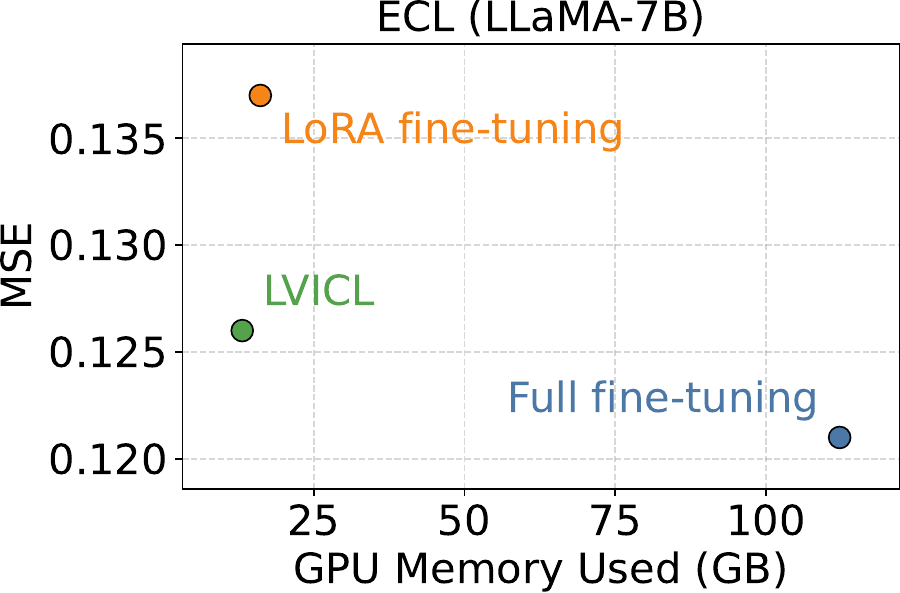}%
    \label{fig_eff1}}
    \hfil
    \subfloat[]{\includegraphics[width=0.30\textwidth]{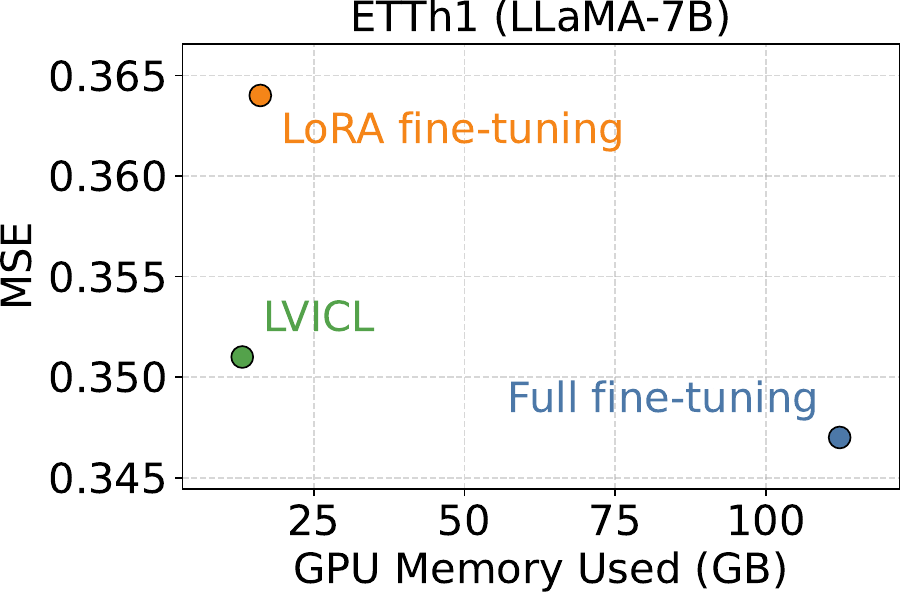}%
    \label{fig_eff2}}
    \hfil
    \subfloat[]{\includegraphics[width=0.30\textwidth]{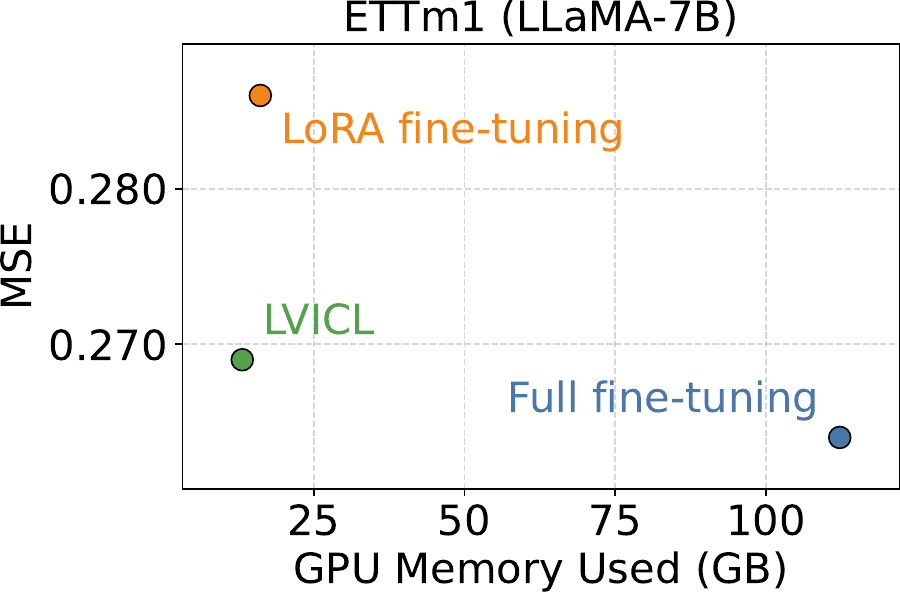}%
    \label{fig_eff3}}
    \hfil
    \caption{Comparison of GPU memory usage and predictive performance for LLMs under different fine-tuning strategies.}
    \Description{}
    \label{fig_eff}
\end{figure*}

\subsection{Zero-shot Forecasting}
To further validate the effectiveness of the proposed method, we compare its predictive performance with other approaches under a zero-shot setting. Specifically, we conduct experiments on the M3 and M4 datasets. The experiments are organized into two groups: 1) train on M4 and test on M3 (M4 → M3), and 2) train on M3 and test on M4 (M3 → M4). For M4 → M3: when the test set is M3 Yearly, M3 Quarterly, and M3 Monthly, the corresponding training sets are M4 Yearly, M4 Quarterly, and M4 Monthly, respectively; when the test set is M3 Others, the training set is M4 Quarterly. For M3 → M4: when the test set is M4 Yearly, M4 Quarterly, and M4 Monthly, the corresponding training sets are M3 Yearly, M3 Quarterly, and M3 Monthly, respectively; when the test sets are M4 Others, the training set is M3 Monthly. The results are shown in Table \ref{tab:tsf_no_multirow}. As can be seen from the table, all results are averages over five runs with different random seeds. LVICL outperforms prior methods under both the M4 → M3 and M3 → M4 settings.

\subsection{Ablation Experiment}
In this section, we conduct ablation experiments. In all experiments, unless otherwise emphasized, the look-back length is fixed at 672, and the prediction length is fixed at 96.

\subsubsection{Post-processing Strategies for Context Vectors.}
In this section, we examine whether it is necessary to post-process the context vectors with a fully connected layer. We evaluate four settings: 1) no post-processing; 2) a fully connected layer; 3) a fully connected layer with ReLU; and 4) a three-layer MLP (no activations between layers). As shown in Figure \ref{fig_mlp}, the single-layer MLP achieves the best forecasting performance. We conjecture that the context vectors produced by the base model already have strong representational power, so a modest linear transformation suffices for further refinement; in contrast, adding stronger nonlinearity or deeper structures (e.g., ReLU or multi-layer MLPs) may introduce unnecessary noise and even hurt performance.

\subsubsection{Injection Location.}
In our approach, the context vector is injected into every layer of LLM. To validate the necessity of this design, we conducted an ablation study comparing training and evaluation results under four injection-location settings: 1) the first quarter of layers, 2) the last quarter of layers, 3) the middle quarter of layers, and 4) all layers. The experimental results are shown in Figure \ref{fig_insert}. As can be seen, injecting the context vector into every layer yields the best performance.

\subsection{Method Analysis}
\subsubsection{Sensitivity to Example Selection and Order}
\label{sec_Sensitivity}
In the Introduction, we posit that the proposed approach reduces the model’s sensitivity to example selection and ordering. To validate this claim, we use the example set in Figure \ref{fig_motivate} to construct context vectors, on which we train and evaluate LVICL that applies these vectors. We then incorporate the results into the data shown in Figure \ref{fig_motivate}. To highlight our method’s effectiveness, we also include the results of standard LVICL in the figure for comparison. The experimental results are presented in Figure \ref{fig_sense}. The results show that our method substantially mitigates the model’s sensitivity to example selection and ordering; moreover, increasing the number of examples used to construct the context vectors further recovers the predictive advantages of fully fine-tuning.

\subsubsection{Method generality}
\label{sec_gen}
To further verify the generality of the proposed method across different LLMs, we port the LVICL learning paradigm to GPT-2 \cite{radford2019language} and OPT \cite{zhang2022opt} and evaluate their predictive performance. The experimental results are shown in Table \ref{tab_gen_res}. As observed from the table, the models’ predictive performance improves as the parameter scale increases.


\subsubsection{Comparison of Different Training Methods}
\label{sec_compare_lora}
This section compares the memory footprint and predictive performance of three fine-tuning strategies on LLaMA-7B: LoRA fine-tuning, full fine-tuning, and our LVICL. In the experiments, the batch size is set to 256 for LoRA and LVICL, while full fine-tuning can only use a batch size of 1 due to its large memory consumption and requires two GPUs to train. All strategies use a single-layer MLP as both the input and output layers. The trainable parameters differ as follows: LoRA updates the input layer, output layer, and LoRA-specific parameters; full fine-tuning updates all parameters; LVICL updates only the input/output layers and the context vector adapter. The results are shown in Figure \ref{fig_eff}. Even with a batch size of 1, full fine-tuning achieves the best predictive performance, but it needs more than 100GB of GPU memory, which severely limits practical deployment. LVICL’s memory footprint is slightly lower than LoRA’s, yields clearly better performance than LoRA, and is slightly below full fine-tuning. Overall, LVICL reproduces most of the predictive advantages of full fine-tuning at a much lower computational cost, offering better cost-effectiveness and practical viability.

\section{Conclusion}
In this paper, we propose \textbf{LVICL}. With all LLM parameters frozen, LVICL recovers most of the gains of full fine-tuning while markedly reducing GPU memory usage. Specifically, in LVICL, we first integrate the LLM with a learnable context-vector adapter to adaptively extract a context vector from multiple examples. We then add this vector to the residual stream of the LLM during the forward pass to improve predictive performance. Experiments show that LVICL significantly outperforms LoRA and approaches full fine-tuning performance with lower training memory, delivering excellent results across multiple real-world datasets.

\section*{Acknowledgements}
The authors would like to thank the anonymous reviewers for their valuable comments. This work is supported by the National Natural Science Foundation of China (No. 62506355).
\clearpage
\newpage
\bibliographystyle{ACM-Reference-Format}
\bibliography{sample-base}

\appendix
\section*{Appendix}
\section{Dataset Descriptions}
\label{Sec_data_des}
We evaluate the proposed LVICL on multiple real‐world datasets spanning several domains:
\begin{itemize}
\item ETT \cite{Informer}: Covers July 2016 to July 2018 with 7 factors related to power transformers. The dataset includes four subsets: ETTh1 and ETTh2 (hourly sampling), and ETTm1 and ETTm2 (15-minute sampling).
\item Weather \cite{liu2023itransformer}: Contains 21 meteorological variables sampled every 10 minutes, collected from the meteorological station of the Max Planck Institute for Biogeochemistry in 2020.
\item ECL \cite{liu2023itransformer}: Electricity consumption data from 321 users, recorded hourly.
\item Traffic \cite{liu2023itransformer}: Road occupancy rates collected from 862 sensors on highways in the San Francisco Bay Area, recorded hourly, covering January 2015 to December 2016.
\item M4 \cite{liu2024autotimes}: A comprehensive dataset spanning various time series across business, finance, economics, and other domains.
\item M3 \cite{liu2024autotimes}: Although smaller than M4, it likewise contains diverse time series from multiple domains and frequencies.
\end{itemize}

We follow the same data processing pipeline and train/validation/test split protocol as in Autotimes \cite{liu2024autotimes} and iTransformer \cite{liu2023itransformer}: the splits are strictly chronological to ensure there is no data leakage. See Table \ref{tab_data} for more details on each dataset.

\begin{table}[htbp]
  \centering
  \caption{Detailed dataset descriptions. \emph{Dim} denotes the variate number. \emph{Dataset Size} denotes the total number of time points in (Train, Validation, Test) splits, respectively. \emph{Prediction Length} denotes the future time points to be predicted. \emph{Frequency} denotes the sampling interval of time points.}
  \label{tab_data}
  \Description{A table listing datasets (ETTh1, Weather, ECL, Traffic, Solar-Energy, M4, M3 variants), their dimensionality, forecast horizons, split sizes, sampling frequency, and domain information.}
  \setlength{\tabcolsep}{10pt}%
  \resizebox{1\linewidth}{!}{%
    \begin{tabular}{l c c c c c}
      \toprule
      Dataset & Dim & Prediction Length & Dataset Size & Frequency & Information \\
      \midrule
      ETTh1, ETTh2 & 7 & \{96, 192, 336, 720\} & (8545, 2881, 2881) & Hourly & Electricity \\
      ETTm1, ETTm2 & 7 & \{96, 192, 336, 720\} & (34465, 11521, 11521) & 15min & Electricity \\
      Weather & 21 & \{96, 192, 336, 720\} & (36792, 5271, 10540) & 10min & Weather \\
      ECL & 321 & \{96, 192, 336, 720\} & (18317, 2633, 5261) & Hourly & Electricity \\
      Traffic & 862 & \{96, 192, 336, 720\} & (12185, 1757, 3509) & Hourly & Transportation \\
      M4-Yearly & 1 & 6 & (23000, 0, 23000) & Yearly & Demographic \\
      M4-Quarterly & 1 & 8 & (24000, 0, 24000) & Quarterly & Finance \\
      M4-Monthly & 1 & 18 & (48000, 0, 48000) & Monthly & Industry \\
      M4-Weekly & 1 & 13 & (359, 0, 359) & Weekly & Macro \\
      M4-Daily & 1 & 14 & (4227, 0, 4227) & Daily & Micro \\
      M4-Hourly & 1 & 48 & (414, 0, 414) & Hourly & Other \\
      M3-Yearly & 1 & 6 & (645, 0, 645) & Yearly & Demographic \\
      M3-Quarterly & 1 & 8 & (756, 0, 756) & Quarterly & Finance \\
      M3-Monthly & 1 & 18 & (1428, 0, 1428) & Monthly & Industry \\
      M3-Others & 1 & 8 & (174, 0, 174) & Weekly & Macro \\
      \bottomrule
    \end{tabular}%
  }
\end{table}

\begin{table}[htbp]
  \centering
  \caption{The Setting of the length of the patch for different Datasets. }
  \label{tab_patch_l}
  \Description{}
  \setlength{\tabcolsep}{10pt}
  \resizebox{0.7\linewidth}{!}{%
    \begin{tabular}{l c}
      \toprule
      Dataset & Patch Length \\
      \midrule
      ETTh1, ETTh2, ETTm1, ETTm2 & 96 \\
      Weather, ECL, Traffic & 96 \\
      M4-Yearly & 6 \\
      M4-Quarterly & 8 \\
      M4-Monthly & 18 \\
      M4-Weekly & 13 \\
      M4-Daily & 14 \\
      M4-Hourly & 48 \\
      M3-Yearly & 6 \\
      M3-Quarterly & 8 \\
      M3-Monthly & 18 \\
      M3-Others & 8 \\
      \bottomrule
    \end{tabular}%
  }
\end{table}


\begin{figure}[htbp]
    \centering
    \subfloat[]{\includegraphics[width=0.23\textwidth]{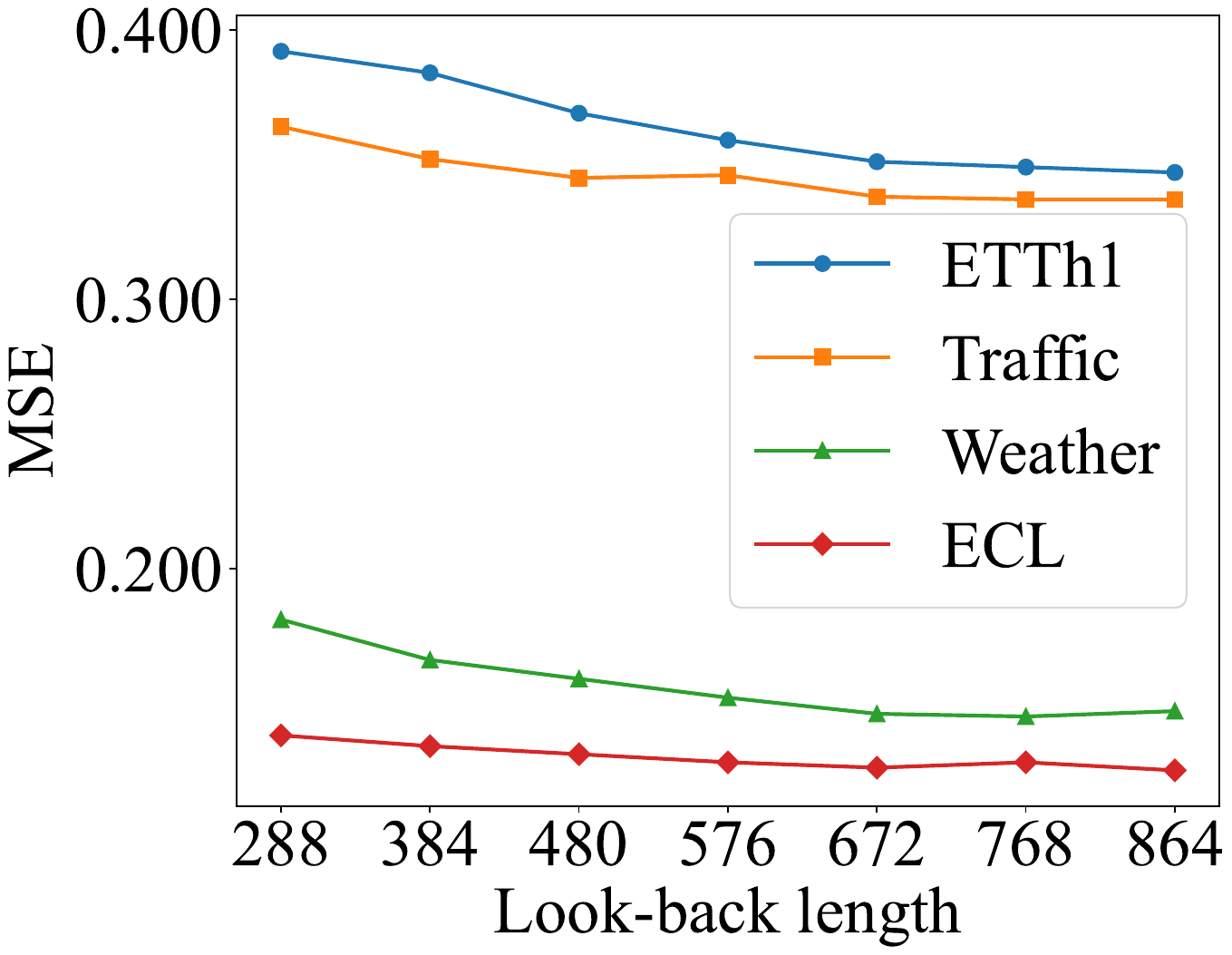}%
    \label{fig_lookback}}
    \hfil
    \subfloat[]{\includegraphics[width=0.23\textwidth]{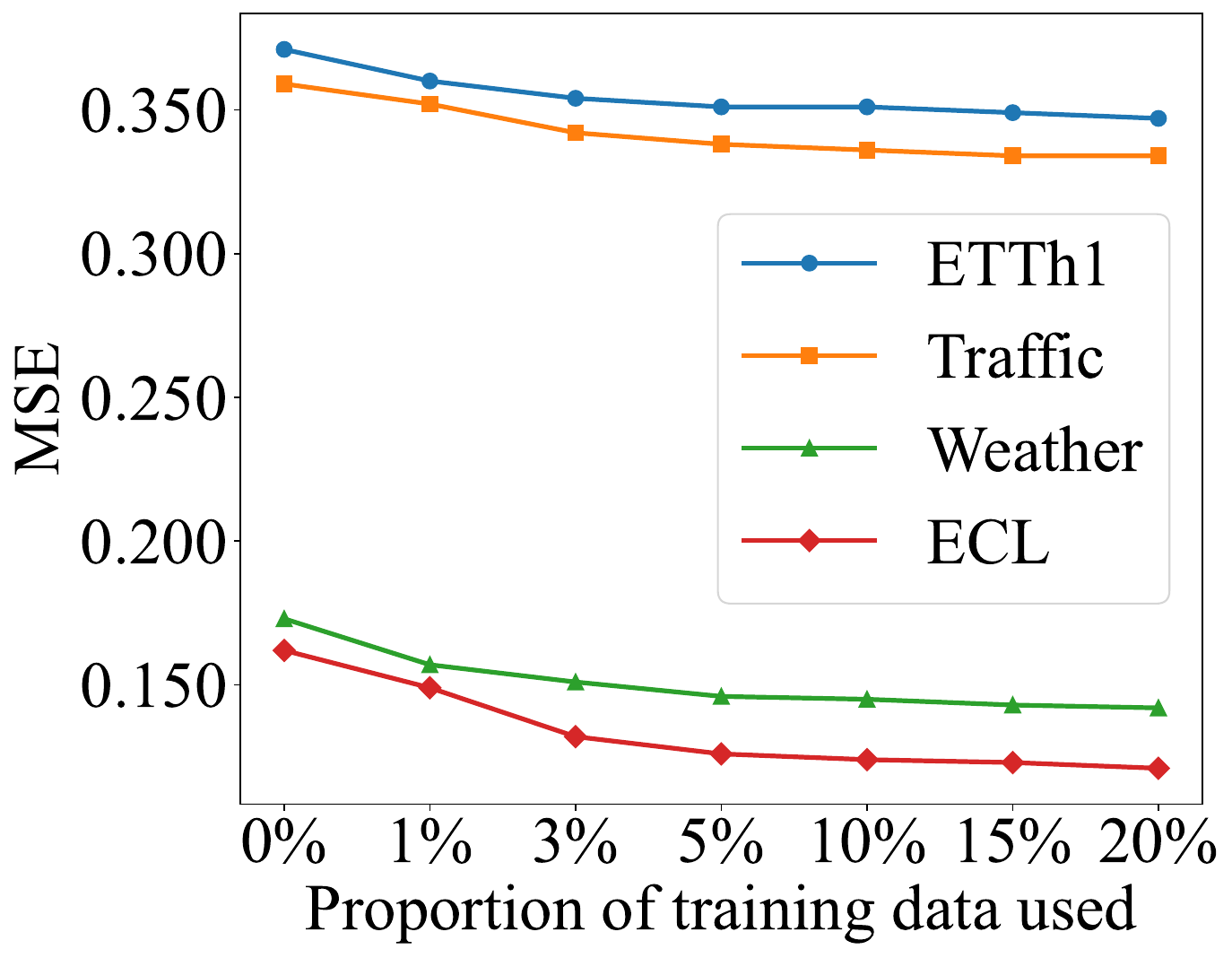}%
    \label{fig_example}}
    \hfil
    \caption{(a) Performance of our method with different look-back lengths. (b) The predictive performance of our method when using different proportions of training data as examples. }
    \Description{}
    \label{fig_combain}
\end{figure}



\begin{figure}[htbp]
    \centering
    \subfloat[]{\includegraphics[width=0.23\textwidth]{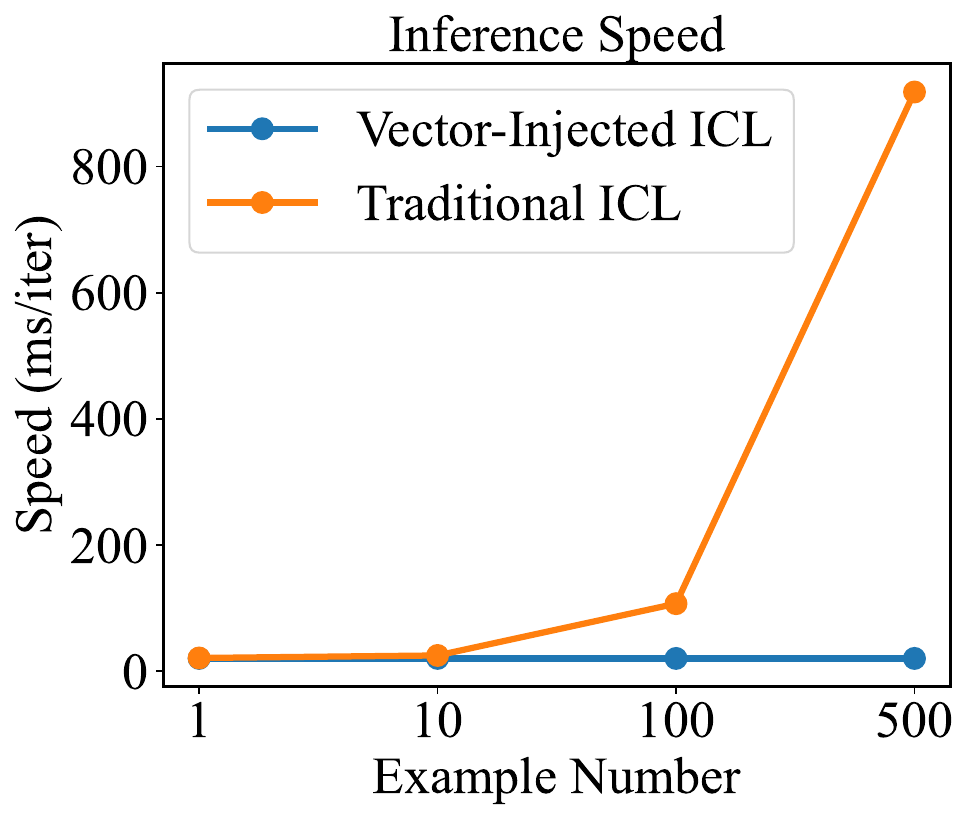}%
    \label{fig_vs1}}
    \hfil
    \subfloat[]{\includegraphics[width=0.23\textwidth]{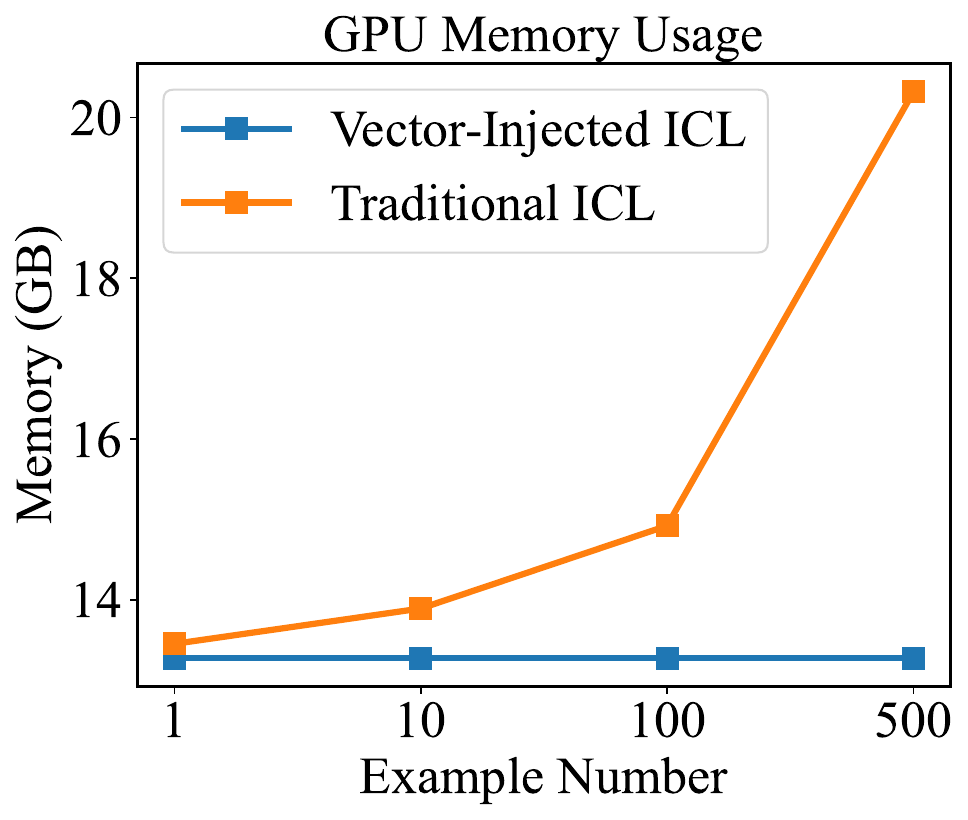}%
    \label{fig_vs2}}
    \hfil
    \caption{Comparison of inference speed and GPU memory between Vector-Injected and Traditional ICL across different numbers of examples.}
    \Description{}
    \label{fig_vs}
\end{figure}

\begin{table*}[!b]
\caption{Full results for the long-term forecasting task. For all prediction horizons, LVICL uses a fixed look-back length of 672. Most results of other methods derive from \cite{liu2024autotimes}. Results not reported in \cite{liu2024autotimes} are obtained by running each method’s source code; the look-back window is selected from \{36, 96, 192, 336, 512, 672, 720\} to report the best results.}
\centering
\small
\setlength{\tabcolsep}{1.5pt} 
\resizebox{1\textwidth}{!}{%
\renewcommand{\arraystretch}{1.4}
\begin{tabular}{ll|rr|rr|rr|rr|rr|rr|rr|rr|rr|rr|rr}
\toprule
\textbf{Dataset} & \textbf{Prediction} & \multicolumn{2}{c}{\textbf{LVICL}} & \multicolumn{2}{c}{AutoTimes} & \multicolumn{2}{c}{TimeLLM} & \multicolumn{2}{c}{FPT} & \multicolumn{2}{c}{Unitime} & \multicolumn{2}{c}{iTransformer} & \multicolumn{2}{c}{DLinear} & \multicolumn{2}{c}{PatchTST} & \multicolumn{2}{c}{TimesNet} & \multicolumn{2}{c}{Chronos} & \multicolumn{2}{c}{TimeFM} \\
 & \textbf{Length} & \textbf{MSE} & \textbf{MAE} & \textbf{MSE} & \textbf{MAE} & \textbf{MSE} & \textbf{MAE} & \textbf{MSE} & \textbf{MAE} & \textbf{MSE} & \textbf{MAE} & \textbf{MSE} & \textbf{MAE} & \textbf{MSE} & \textbf{MAE} & \textbf{MSE} & \textbf{MAE} & \textbf{MSE} & \textbf{MAE} & \textbf{MSE} & \textbf{MAE} & \textbf{MSE} & \textbf{MAE} \\
\midrule
\textbf{ETTh1} & 96 & \textbf{0.351} & \textbf{0.389} & \underline{0.360} & 0.400 & 0.362 & \underline{0.392} & 0.376 & 0.397 & 0.397 & 0.418 & 0.386 & 0.405 & 0.375 & 0.399 & 0.370 & 0.399 & 0.384 & 0.402 & 0.369 & 0.409 & 0.384 & 0.426 \\
 & 192 & \textbf{0.379} & \textbf{0.408} & \underline{0.388} & 0.419 & 0.398 & 0.418 & 0.416 & 0.418 & 0.434 & 0.439 & 0.422 & 0.439 & 0.405 & \underline{0.416} & 0.413 & 0.421 & 0.557 & 0.436 & 0.399 & 0.429 & 0.415 & 0.447 \\
 & 336 & \textbf{0.392} & \textbf{0.417} & \underline{0.401} & 0.429 & 0.430 & \underline{0.427} & 0.442 & 0.433 & 0.468 & 0.458 & 0.444 & 0.457 & 0.439 & 0.443 & 0.422 & 0.436 & 0.491 & 0.469 & 0.413 & 0.439 & 0.429 & 0.456 \\
 & 720 & \textbf{0.402} & \textbf{0.434} & \underline{0.406} & \underline{0.440} & 0.442 & 0.457 & 0.477 & 0.456 & 0.469 & 0.477 & 0.500 & 0.498 & 0.472 & 0.490 & 0.447 & 0.466 & 0.521 & 0.500 & 0.423 & 0.457 & 0.440 & 0.475 \\
 & \textbf{Avg} & \textbf{0.381} & \textbf{0.412} & \underline{0.389} & \underline{0.422} & 0.408 & 0.423 & 0.427 & 0.426 & 0.442 & 0.448 & 0.438 & 0.450 & 0.423 & 0.437 & 0.413 & 0.431 & 0.458 & 0.450 & 0.401 & 0.434 & 0.417 & 0.451 \\
\midrule
\textbf{ETTh2} & 96 & \textbf{0.270} & \textbf{0.332} & 0.282 & 0.342 & 0.288 & 0.341 & 0.287 & 0.341 & 0.297 & 0.354 & 0.304 & 0.360 & 0.289 & 0.353 & 0.274 & 0.336 & 0.340 & 0.374 & \underline{0.271} & \underline{0.333} & 0.277 & 0.340 \\
 & 192 & \underline{0.333} & \underline{0.378} & 0.348 & 0.387 & 0.351 & 0.389 & 0.350 & 0.383 & 0.365 & 0.404 & 0.377 & 0.403 & 0.383 & 0.418 & 0.339 & 0.379 & 0.402 & 0.414 & \textbf{0.326} & \textbf{0.371} & 0.334 & 0.380 \\
 & 336 & \textbf{0.325} & \textbf{0.377} & 0.365 & 0.412 & 0.362 & 0.401 & 0.373 & 0.395 & 0.359 & 0.401 & 0.405 & 0.429 & 0.448 & 0.465 & \underline{0.329} & \underline{0.380} & 0.452 & 0.452 & 0.338 & 0.394 & 0.340 & 0.393 \\
 & 720 & \underline{0.375} & \underline{0.416} & 0.412 & 0.440 & 0.415 & 0.440 & 0.401 & 0.443 & 0.403 & 0.436 & 0.443 & 0.464 & 0.605 & 0.551 & 0.379 & 0.422 & 0.462 & 0.468 & \textbf{0.372} & \textbf{0.409} & 0.384 & 0.426 \\
 & \textbf{Avg} & \textbf{0.326} & \textbf{0.376} & 0.352 & 0.395 & 0.354 & 0.393 & 0.353 & 0.391 & 0.356 & 0.399 & 0.382 & 0.414 & 0.431 & 0.446 & 0.330 & 0.379 & 0.414 & 0.427 & \underline{0.327} & \underline{0.377} & 0.334 & 0.385 \\
\midrule
\textbf{ETTm1} & 96 & \underline{0.269} & 0.343 & 0.274 & 0.343 & 0.284 & \underline{0.341} & 0.301 & 0.343 & 0.308 & 0.358 & 0.312 & 0.366 & 0.299 & 0.343 & 0.290 & 0.342 & 0.338 & 0.375 & \textbf{0.264} & \textbf{0.337} & 0.280 & 0.357 \\
 & 192 & \underline{0.309} & 0.371 & 0.316 & 0.370 & 0.327 & \underline{0.363} & 0.348 & \textbf{0.362} & 0.354 & 0.391 & 0.347 & 0.385 & 0.335 & 0.365 & 0.332 & 0.369 & 0.374 & 0.387 & \textbf{0.303} & 0.364 & 0.321 & 0.386 \\
 & 336 & \underline{0.343} & 0.390 & 0.344 & 0.390 & 0.368 & 0.387 & 0.386 & 0.397 & 0.396 & 0.413 & 0.379 & 0.404 & 0.369 & \underline{0.386} & 0.366 & 0.392 & 0.410 & 0.411 & \textbf{0.337} & \textbf{0.383} & 0.357 & 0.405 \\
 & 720 & \underline{0.391} & \underline{0.408} & 0.392 & 0.418 & 0.423 & 0.419 & 0.430 & 0.427 & 0.442 & 0.450 & 0.441 & 0.442 & 0.425 & 0.421 & 0.416 & 0.420 & 0.478 & 0.450 & \textbf{0.384} & \textbf{0.401} & 0.406 & 0.424 \\
 & \textbf{Avg} & \underline{0.328} & \underline{0.378} & 0.332 & 0.380 & 0.350 & \underline{0.378} & 0.366 & 0.382 & 0.375 & 0.403 & 0.370 & 0.399 & 0.357 & \underline{0.378} & 0.351 & 0.380 & 0.400 & 0.406 & \textbf{0.322} & \textbf{0.371} & 0.341 & 0.393 \\
\midrule
\textbf{ETTm2} & 96 & \textbf{0.155} & \textbf{0.246} & 0.159 & 0.250 & 0.168 & 0.250 & 0.172 & 0.258 & 0.179 & 0.265 & 0.179 & 0.271 & 0.167 & 0.269 & 0.165 & 0.255 & 0.187 & 0.267 & 0.159 & 0.252 & \underline{0.157} & \underline{0.249} \\
 & 192 & \textbf{0.210} & \textbf{0.282} & 0.213 & \underline{0.285} & 0.216 & 0.294 & 0.231 & 0.287 & 0.239 & 0.309 & 0.242 & 0.313 & 0.224 & 0.303 & 0.220 & 0.292 & 0.249 & 0.309 & 0.215 & 0.288 & \underline{0.212} & \underline{0.285} \\
 & 336 & \textbf{0.258} & \textbf{0.317} & \underline{0.261} & 0.323 & 0.269 & 0.323 & 0.282 & 0.329 & 0.293 & 0.346 & 0.288 & 0.344 & 0.281 & 0.342 & 0.274 & 0.329 & 0.321 & 0.351 & 0.264 & 0.324 & \underline{0.261} & \underline{0.320} \\
 & 720 & \textbf{0.335} & \textbf{0.378} & 0.341 & \underline{0.381} & 0.363 & 0.387 & 0.373 & 0.387 & 0.398 & 0.397 & 0.378 & 0.397 & 0.397 & 0.421 & 0.362 & 0.385 & 0.408 & 0.403 & 0.343 & 0.387 & \underline{0.338} & 0.382 \\
 & \textbf{Avg} & \textbf{0.239} & \textbf{0.306} & 0.243 & 0.310 & 0.254 & 0.314 & 0.265 & 0.315 & 0.277 & 0.329 & 0.272 & 0.331 & 0.267 & 0.333 & 0.255 & 0.315 & 0.291 & 0.333 & 0.245 & 0.313 & \underline{0.242} & \underline{0.309} \\
\midrule
\textbf{ECL} & 96 & \textbf{0.126} & \underline{0.224} & \underline{0.129} & 0.225 & 0.131 & \underline{0.224} & 0.139 & 0.238 & 0.196 & 0.287 & 0.132 & 0.227 & 0.153 & 0.237 & \underline{0.129} & \textbf{0.222} & 0.168 & 0.272 & 0.131 & 0.233 & \underline{0.129} & 0.230 \\
 & 192 & \underline{0.148} & \textbf{0.235} & \textbf{0.147} & 0.241 & 0.152 & 0.241 & 0.153 & 0.251 & 0.199 & 0.291 & 0.153 & 0.249 & 0.152 & 0.249 & \textbf{0.147} & \underline{0.240} & 0.184 & 0.289 & 0.154 & 0.244 & 0.152 & 0.241 \\
 & 336 & \underline{0.161} & \underline{0.252} & 0.162 & 0.258 & \textbf{0.160} & \textbf{0.248} & 0.169 & 0.266 & 0.214 & 0.305 & 0.167 & 0.262 & 0.169 & 0.267 & 0.163 & 0.259 & 0.198 & 0.300 & 0.167 & 0.262 & 0.165 & 0.259 \\
 & 720 & \underline{0.196} & \textbf{0.282} & 0.199 & \underline{0.288} & \textbf{0.192} & 0.298 & 0.206 & 0.297 & 0.254 & 0.335 & 0.254 & 0.335 & 0.233 & 0.344 & 0.197 & 0.290 & 0.220 & 0.320 & 0.204 & 0.293 & 0.201 & 0.290 \\
 & \textbf{Avg} & \textbf{0.158} & \textbf{0.248} & \underline{0.159} & \underline{0.253} & \underline{0.159} & \underline{0.253} & 0.167 & 0.263 & 0.216 & 0.305 & 0.161 & 0.256 & 0.177 & 0.274 & \underline{0.159} & \underline{0.253} & 0.192 & 0.295 & 0.164 & 0.258 & 0.162 & 0.255 \\
\midrule
\textbf{Weather} & 96 & \underline{0.146} & \textbf{0.196} & 0.153 & 0.203 & 0.147 & 0.201 & 0.162 & 0.212 & 0.171 & 0.214 & 0.163 & 0.211 & 0.152 & 0.237 & 0.149 & \underline{0.198} & 0.172 & 0.220 & \textbf{0.145} & \underline{0.198} & 0.151 & 0.202 \\
 & 192 & 0.190 & \underline{0.240} & 0.201 & 0.250 & \underline{0.189} & \textbf{0.234} & 0.204 & 0.248 & 0.217 & 0.254 & 0.205 & 0.250 & 0.220 & 0.282 & 0.194 & 0.241 & 0.219 & 0.261 & \textbf{0.188} & 0.242 & 0.196 & 0.248 \\
 & 336 & \underline{0.240} & \underline{0.280} & 0.256 & 0.293 & 0.262 & \textbf{0.279} & 0.254 & 0.286 & 0.274 & 0.293 & 0.254 & 0.289 & 0.265 & 0.319 & 0.245 & 0.282 & 0.280 & 0.306 & \textbf{0.238} & 0.281 & 0.248 & 0.289 \\
 & 720 & 0.312 & \underline{0.330} & 0.331 & 0.345 & \textbf{0.304} & \textbf{0.316} & 0.326 & 0.337 & 0.329 & 0.340 & 0.329 & 0.340 & 0.323 & 0.362 & 0.314 & 0.334 & 0.365 & 0.359 & \underline{0.309} & 0.331 & 0.322 & 0.340 \\
 & \textbf{Avg} & \underline{0.224} & \underline{0.262} & 0.235 & 0.273 & 0.225 & \textbf{0.257} & 0.237 & 0.270 & 0.253 & 0.276 & 0.238 & 0.272 & 0.240 & 0.300 & 0.226 & 0.264 & 0.259 & 0.287 & \textbf{0.220} & 0.263 & 0.229 & 0.270 \\
\midrule
\textbf{Traffic} & 96 & \textbf{0.338} & \textbf{0.245} & \underline{0.343} & \underline{0.248} & 0.362 & \underline{0.248} & 0.388 & 0.282 & 0.401 & 0.291 & 0.351 & 0.257 & 0.410 & 0.282 & 0.360 & 0.249 & 0.593 & 0.321 & \underline{0.343} & 0.249 & 0.350 & 0.253 \\
 & 192 & \textbf{0.360} & \underline{0.251} & \underline{0.362} & 0.257 & 0.374 & \textbf{0.247} & 0.407 & 0.290 & 0.432 & 0.298 & 0.364 & 0.265 & 0.423 & 0.287 & 0.379 & 0.256 & 0.617 & 0.333 & 0.366 & 0.255 & 0.372 & 0.260 \\
 & 336 & \underline{0.382} & \textbf{0.259} & \textbf{0.379} & 0.266 & 0.385 & 0.271 & 0.412 & 0.294 & 0.436 & 0.305 & \underline{0.382} & 0.273 & 0.436 & 0.296 & 0.392 & 0.264 & 0.629 & 0.333 & 0.388 & \underline{0.263} & 0.395 & 0.268 \\
 & 720 & \textbf{0.401} & \underline{0.285} & 0.413 & \textbf{0.284} & 0.430 & 0.288 & 0.450 & 0.312 & 0.447 & 0.322 & 0.420 & 0.292 & 0.466 & 0.315 & 0.432 & 0.286 & 0.640 & 0.330 & \underline{0.407} & 0.289 & 0.415 & 0.295 \\
 & \textbf{Avg} & \textbf{0.370} & \textbf{0.260} & \underline{0.374} & \underline{0.264} & 0.388 & \underline{0.264} & 0.414 & 0.294 & 0.429 & 0.304 & 0.379 & 0.272 & 0.434 & 0.295 & 0.391 & \underline{0.264} & 0.620 & 0.336 & 0.376 & \underline{0.264} & 0.383 & 0.269 \\
\bottomrule
\end{tabular}
}
\label{tab_long_res}
\end{table*}

\section{More Implementation Details}
For different datasets, we use different patch lengths. The specific settings are shown in Table \ref{tab_patch_l}. The effectiveness of this configuration is discussed in detail in \cite{liu2024autotimes}.

\section{Hyperparameter Sensitivity Analysis}
\label{sec_hsa}
To select the optimal parameter configuration, in this subsection, we conduct a sensitivity analysis of three key hyperparameters: the look-back length and the number of examples used to construct the context vector. 

\subsection{Variable look-back length}
We systematically evaluate our method under multiple look-back length configurations to examine its sensitivity and robustness to changes in the historical window. In all experiments, the prediction length is set to 96. The results are shown in Figure \ref{fig_lookback}: although the metrics fluctuate across settings, the overall trend is clear— as the look-back length increases, the model leverages richer historical context and achieves steady improvements in predictive accuracy. However, excessively long windows slow inference, and the performance gains taper off once the look-back length exceeds 672. Balancing predictive benefits against computational cost, we set the look-back length to 672.

\subsection{Number of Examples Used}
\label{sec_example_num}
In this section, we evaluate how using different proportions of the training data (0\%–20\%) as examples affects the model’s forecasting performance. The experimental setup fixes the look-back window at 672 and the forecast horizon at 96; results are shown in Fig. \ref{fig_example}. We observe that 1) performance improves as the proportion of training data used for examples increases, and 2) once the proportion exceeds 5\%, the growth in gains slows noticeably. Accordingly, we use 5\% of the training data as examples to construct the context vector, striking a balance between performance and efficiency. Note that all examples are compressed into a single context vector, so the number of examples does not affect the model’s inference speed.

\section{Inference Efficiency Comparison of Vector-Injected and Traditional ICL}
\label{sec_IE_ICL}
To further highlight the inference efficiency advantages of Vector-Injected ICL, we compare the GPU memory usage and inference speed of the two ICL methods across varying numbers of examples; see Fig. \ref{fig_vs}. The results show that as the number of examples increases, the GPU memory usage and inference speed of traditional ICL rise markedly. In contrast, Vector-Injected ICL remains nearly unchanged on both metrics. This indicates that Vector-Injected ICL makes injecting more examples feasible while keeping computational and memory overheads under control.

\section{More Method Analysis}
\subsection{Effectiveness of Aggregation}
\label{sec_Aggregation}
In LVICL, we aggregate feature vectors from multiple samples into a single context vector to capture features shared across samples. We hypothesize that such shared features are more relevant to the prediction task than single-sample representations. To validate this, we measure the average mutual information (MI) between the aggregated context vector constructed from varying numbers of samples from 1 to 5000 and the predicted series. There is no overlap between the predicted series and the samples participating in the aggregation. MI quantifies statistical dependence, with higher values indicating stronger correlation. The predicted series (length $P$) is transformed via a trained embedding layer to match the context vector's dimensions, after which their MI is computed using the histogram method. Further details on the histogram method are in \cite{steuer2002mutual}. The results are shown in Figure \ref{fig_Aggregation_MI}, where each point averages 20 random trials. The results show that both mean and adaptive weighted aggregation yield significantly higher MI than single-sample representations, and the MI increases with the number of aggregated samples. These results indicate that cross-sample shared features are more strongly correlated with the prediction task.

\subsection{Comparison of Aggregation Methods}
\label{sec_Aggregation_method}
\begin{figure}[htbp]
    \centering
    \subfloat[]{\includegraphics[width=0.20\textwidth]{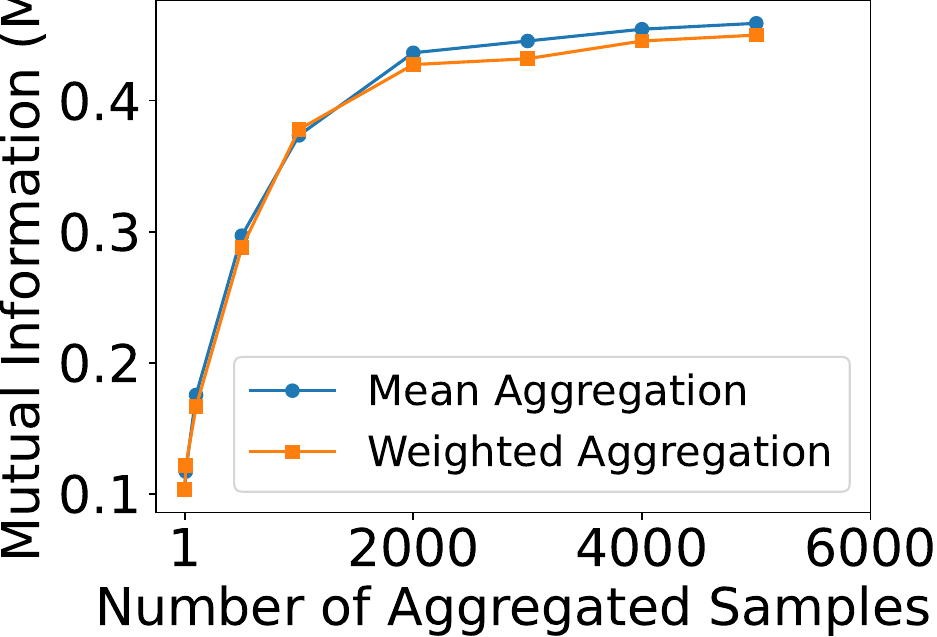}%
    \label{fig_Aggregation_MI}}
    \hfil
    \subfloat[]{\includegraphics[width=0.23\textwidth]{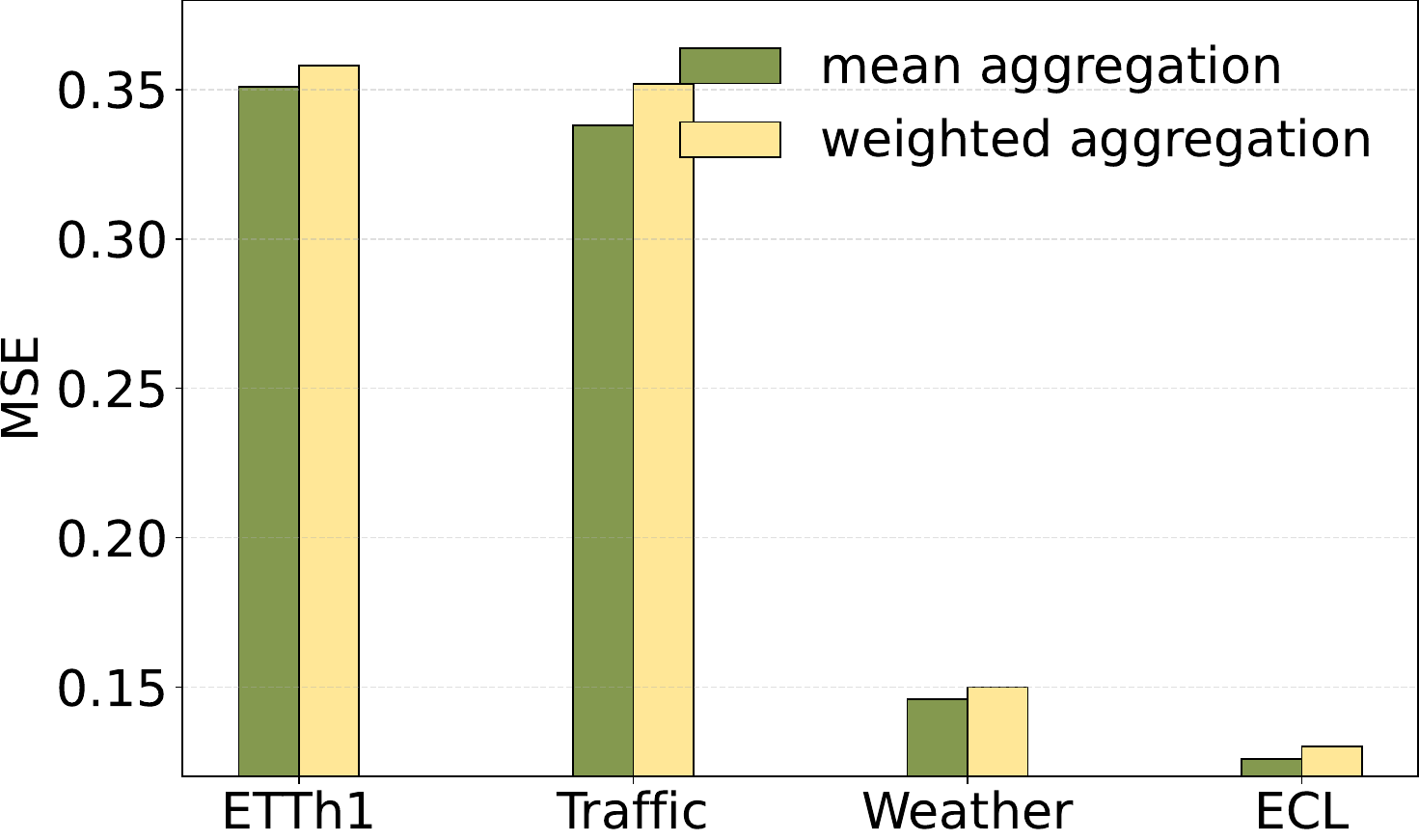}%
    \label{fig_Aggregation}}
    \hfil
    \caption{(a) Average MI between the aggregated context vector and the predicted series
under different numbers of aggregated samples. (b) Comparison of two aggregation methods.}
    \Description{}
    \label{fig_temp}
\end{figure}

We compare the final performance of the model using mean aggregation in Eq. \ref{eq_eh_e} and adaptive weighted aggregation. The experimental results are shown in Figure \ref{fig_Aggregation}. The results indicate that mean aggregation outperforms adaptive weighted aggregation.

\subsection{Effectiveness of Context Vector Adapter}
\label{sec_Adapter}
We compute the MI between the context vector and the predicted series, both before and after processing by the Context Vector Adapter. The results show that the values are comparable, remaining at approximately 0.56. However, the ablation study in Figure \ref{fig_mlp} demonstrates that using the context vector processed by the adapter yields a significantly lower prediction MSE. This indicates that the adapter effectively filters out prediction-irrelevant redundant information from the context vector while preserving key prediction-related features, thereby enhancing the model's overall performance.

\section{Full Forecasting Results}
\label{full_res}
The full multivariate forecasting results are provided in the section due to the space limitation of the main text (Table \ref{tab_long_res}). We extensively evaluate competitive counterparts on challenging forecasting tasks. All results are averaged over five random seeds.

\end{document}